\documentclass{article}
\usepackage{geometry}
\usepackage[utf8]{inputenc}
\usepackage{cite}
\usepackage{url}
\usepackage{amssymb,amsmath,amsthm}
\usepackage{bm}
\usepackage{graphicx}
\usepackage{enumerate}
\usepackage{microtype}
\usepackage{color}
\usepackage{hyperref}
\usepackage{multirow}
\usepackage{svg}
\usepackage{array}
\usepackage{float}
\usepackage{listings}
\usepackage{subcaption}
\usepackage{booktabs}
\usepackage[dvipsnames]{xcolor}
\newcolumntype{C}[1]{>{\centering\let\newline\\\arraybackslash\hspace{0pt}}m{#1}}
\graphicspath{{Figures/}}

\usepackage[ruled,linesnumbered,vlined]{algorithm2e}

\title{Resolution-free Neural Surrogates for Geometric Parameterization and Mapping with Spatially Varying Fields}

\author{Yanwen Huang\thanks{Department of Mathematics, The Chinese University of Hong Kong
  ({ywhuang@math.cuhk.edu.hk}).}
\and Lok Ming Lui\thanks{Department of Mathematics, The Chinese University of Hong Kong
  ({lmlui@math.cuhk.edu.hk}).}
\and Gary P. T. Choi\thanks{Department of Mathematics, The Chinese University of Hong Kong
  ({ptchoi@cuhk.edu.hk}).}}

\date{ }

\begin{document}

\maketitle

\begin{abstract}
Many imaging problems require computing spatial transformations induced by spatially varying intensity, feature, or density fields. Canonical examples include distortion correction, deformable image registration, atlas-based segmentation, and deformation-driven image analysis. These tasks can be formulated as geometric mapping problems in which the transformation is constrained to preserve local structure, control boundary behavior, or regulate angular distortion. Such formulations typically lead to variational models, diffusion processes, or elliptic partial differential equations. However, repeatedly solving high-resolution systems becomes computationally expensive when the underlying parameter fields vary across instances. In this work, we propose a resolution-free neural surrogate for geometric parameterization and mapping problems. Given a spatially varying parameter field $p:\Omega\to\mathbb{R}^m$ and grid nodes $\{x_i\}_{i=1}^N\subset\Omega$, the model predicts mapped locations $\{u(x_i)\}_{i=1}^N$ on structured grids of different sizes without retraining. To avoid dependence on a fixed grid, we use a multi-resolution geometric encoding strategy that conditions the network on coordinate-augmented samples of the parameter field. The model is trained without labeled solution data by enforcing geometry-aware constraints derived from variational energies, diffusion-based density equalization, and quasi-conformal theory. Experimental results on quasi-conformal mapping and density-equalizing mapping problems are presented to demonstrate the effectiveness of our proposed method.
\end{abstract}

\section{Introduction}
\label{sect:intro}

Geometric parameterization and mapping problems are fundamental in geometry processing, scientific visualization, computational anatomy, mesh deformation, and imaging applications~\cite{Floater2005surface,Sheffer2007mesh}. Imaging problems such as distortion correction, deformable registration, atlas construction, and density-guided visualization can be viewed as transformations of image domains driven by intensity, feature, or density information, with the goal of preserving spatial coherence, boundary consistency, and local geometric structures~\cite{harten2023deformable,choi2018density}. The main objective is to construct a mapping $u:\Omega\to\mathbb{R}^d$ that transforms points in a source domain into a target configuration while preserving or controlling important geometric structures. Depending on the application, the mapping may be required to preserve boundary correspondence, reduce angular distortion, redistribute density, maintain local injectivity, or produce a smooth deformation field.

Unlike ordinary pointwise prediction problems, geometric mapping is inherently relational. The image of a point cannot be determined independently of its surrounding points, because local neighborhoods encode geometric information such as distances, orientations, areas, angles, and connectivity. For example, in a parameterization problem, nearby points in the original domain should usually remain nearby after mapping so that the local structure of the domain is not destroyed. In quasi-conformal mapping, the deformation of an infinitesimal neighborhood is controlled by the Beltrami coefficient, which describes local angular distortion. In density-equalizing mapping, the displacement of each point is coupled with the surrounding density distribution, since the mapping must redistribute mass over the entire domain. Therefore, the desired mapping should be understood not only as a pointwise function, but also as a transformation that preserves and modifies neighborhood relations in a geometrically meaningful way.

Many geometric mapping problems are driven by spatially varying parameter fields. Let $ p:\Omega\to\mathbb{R}^m$ denote such a field. Depending on the task, $p$ may represent a density function, a Beltrami coefficient, a metric tensor, a material descriptor, or another geometric quantity. The mapping $u$ is expected to respond to this field while maintaining consistency with both local and global geometric constraints. For instance, a density field determines how regions should expand or shrink in a density-equalizing map, while a Beltrami coefficient specifies the local distortion pattern of a quasi-conformal parameterization.

Classical numerical methods usually construct these mappings by discretizing the domain and solving a system derived from a geometric model. Such models may be formulated through differential constraints, variational principles, or diffusion processes. For example, quasi-conformal maps can be characterized through the Beltrami equation~\cite{ahlfors2006lectures,choi2022recent}, while density-equalizing maps can be derived from diffusion equations and associated velocity fields~\cite{gastner2004diffusion,choi2018density}. Variational formulations are also widely used, where the desired mapping is obtained by minimizing an energy functional that penalizes distortion, density mismatch, or violation of geometric constraints. However, many classical algorithms are designed for a specific type of domain, such as a planar region, an open surface, a spherical surface, or a volumetric domain. Extending the method to a new geometry or topology often requires substantial algorithmic modification~\cite{lyu2024bijective,lyu2024spherical,lyu2026ellipsoidal,choi2021volumetric}.

In many practical settings, the parameter field changes repeatedly across subjects, time frames, simulations, or design iterations. Each new parameter field may require recomputing a new geometric map. When high spatial resolution is needed, repeated numerical solution can become expensive, especially for nonlinear geometric energies or mappings with strong local deformation. This motivates the development of learned surrogate models that approximate the mapping operator
$\mathcal{G}:p(\cdot)\mapsto u(\cdot),$
so that a trained model can rapidly predict the mapping associated with a new parameter field~\cite{lu2021learning,li2021fourier}. However, learning such an operator is challenging for geometric parameterization and mapping problems. First, the input parameter is a continuous field rather than a finite-dimensional vector. A direct grid-based representation ties the model to a fixed input resolution and makes it difficult to transfer across different discretizations~\cite{lee2022mesh,roy2025pimrno}. Second, the output mapping should be evaluable at arbitrary spatial locations, since different applications may require different point sets, grid sizes, or mesh resolutions. Third, the model must respect neighborhood-dependent geometric structures. A valid mapping should not only predict the displacement of individual points, but should also preserve local consistency among neighboring points and avoid undesirable artifacts such as folding, excessive distortion, or boundary violation~\cite{lyu2024bijective}.

Recent neural-operator methods provide a promising direction for learning mappings between function spaces. Physics-informed neural operators further reduce the need for labeled training pairs by enforcing governing equations or residual constraints during training \cite{zhu2019physics}. Related data-free multi-resolution neural-operator approaches have shown that physical constraints can be imposed directly on discretized fields \cite{roy2025pimrno}. Nevertheless, many existing methods remain tied to fixed input sampling, even when their output representation is resolution-independent. This limitation is especially restrictive for geometric mapping problems, where both the parameter field and the evaluation points may vary in resolution or be sampled on unstructured point sets.

\begin{figure}[t]
    \centering
    \includegraphics[width=1.0\linewidth]{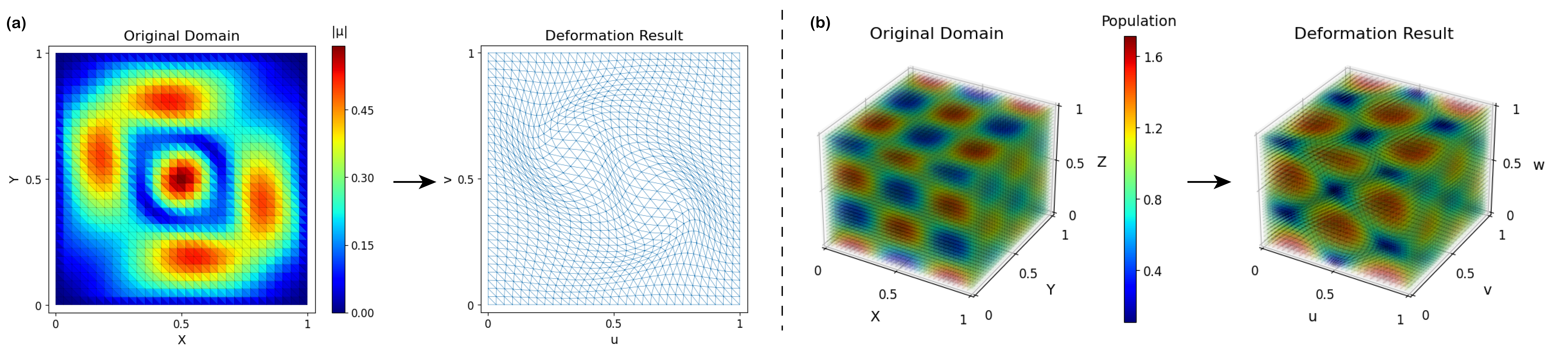}
    \caption{\textbf{Examples of geometric parameterizations and mappings achieved by our proposed resolution-free neural surrogate.} (a) Given a Beltrami coefficient field defined on a 2D domain, our framework produces a quasi-conformal parameterization with the desired local geometric changes. (b) Given a density distribution in a 3D domain, our framework produces a density-equalizing map with the desired local volume changes.}
    \label{fig:illustration}
\end{figure}

To address these issues, we develop a neural surrogate for geometric mapping on structured grids of different sizes (see Fig.~\ref{fig:illustration} for examples). The model combines sampled coordinates and parameter values with multiscale features. It predicts a displacement at each input grid node, and the final map is obtained by deforming the reference coordinates. Our resolution-free framework is able to reuse the same trained weights across the supported grid sizes.

The proposed framework is trained without paired ground-truth solution data. Instead, training is performed by minimizing geometry-aware objectives derived from the mathematical structure of the mapping problem. For quasi-conformal parameterization, the loss enforces consistency with Beltrami-type distortion constraints. For density-equalizing maps, the loss encourages the mapping to redistribute density consistently with diffusion-based or variational principles. These objectives guide the network to learn geometrically meaningful mappings while avoiding the cost of generating large supervised datasets from classical solvers.

Our contributions are as follows:
\begin{enumerate}
\item \textbf{Resolution-free surrogate for geometric parameterization and mapping.}
We propose a neural surrogate that learns mappings from spatial parameter fields to geometric deformation fields. The model can evaluate mappings on structured grids of different sizes without retraining.

\item \textbf{Neighborhood-aware geometric field representation.}
We encode coordinates and spatial parameter fields in a unified representation so that the network can learn how local neighborhoods, spatial relations, and geometric context influence the mapped positions.

\item \textbf{Data-free training with geometry-aware constraints.}
The surrogate is trained without paired numerical solutions by enforcing constraints derived from variational principles, diffusion-based density equalization, and quasi-conformal theory.

\item \textbf{Applications to representative mapping problems.}
We validate the method on quasi-conformal mapping governed by the Beltrami equation and density-equalizing mapping problems formulated through diffusion-driven and variational objectives.
\end{enumerate}

The remainder of the paper is organized as follows. Section~\ref{sect:previous} reviews related work on operator learning, physics-constrained neural surrogates, and geometric mapping methods. Section~\ref{sect:background} summarizes the mathematical background used in this work, including quasi-conformal theory and diffusion-based density-equalizing maps. Section~\ref{sect:method} presents the proposed resolution-free architecture, geometric data encoding strategy, model configuration, and weight refinement procedure. Section~\ref{sect:experiment} reports experiments on quasi-conformal parameterization and density-equalizing mapping problems and demonstrates the effectiveness of our model for handling various problems in both 2D and 3D. We also compare our method with the Learning-based Density-Equalizing Mapping (LDEM) approach~\cite{huang2025ldem} and report an ablation of the weight refinement procedure. We conclude the paper and discuss possible directions in Section~\ref{sect:conclusion}.

\section{Previous Work}
\label{sect:previous}

Geometric parameterization and mapping problems aim to construct transformations between geometric domains while preserving important structural properties, such as local neighborhood relationships, boundary conditions, angular structure, and density distributions. These problems arise in many areas, including computer graphics, geometry processing, and scientific visualization. However, they remain challenging when the underlying parameter fields vary across different problem instances. Although traditional numerical methods can provide accurate solutions, they often become computationally expensive at high resolutions or when many repeated computations are needed. Neural surrogate models offer a promising alternative by learning an approximation to the mapping operator itself. This enables efficient inference across different spatial resolutions and reduces the computational cost associated with classical iterative solvers.

\subsection{Traditional Mapping and Parameterization Methods}
Traditional geometric mapping relies on mesh discretization to solve systems derived from differential geometry~\cite{Floater2005surface,gu2023classical}. Surface parameterization methods commonly aim to minimize angular distortion, area distortion, or local shape distortion, which are important for texture mapping, remeshing, and geometric modeling~\cite{Sheffer2007mesh}. Classical techniques such as Tutte’s graph embedding and harmonic or barycentric mappings formulate the problem as sparse linear systems\cite{Floater1997param}, while methods such as Angle-Based Flattening (ABF)~\cite{Sheffer2001abf} and Least-Squares Conformal Maps (LSCM)~\cite{Levy2002lscm} introduce nonlinear or least-squares optimization to improve conformality. Besides angle preservation, many other approaches have also considered the preservation of area or the balance between angle and area, as well as their higher-dimensional extensions~\cite{wang2016arap++,yueh2019novel}.

For density-equalizing maps (DEM), related cartogram methods deform spatial domains so that the area reflects a prescribed density distribution. Early continuous cartogram algorithms used iterative displacement or force-based deformation~\cite{Dougenik1985cartograms}, while later methods such as CartoDraw~\cite{Keim2004cartodraw} and diffusion-based cartograms~\cite{gastner2004diffusion,choi2018density} improved shape preservation and density redistribution through scanline heuristics or physical diffusion models. However, these classical solvers can be computationally intensive at high resolutions and may produce mesh overlaps, foldovers, or artifacts in regions with extreme parameter gradients~\cite{huang2025ldem,lyu2024bijective}.

\subsection{Evolution of Neural Operators and Resolution Independence}

Traditional deep learning models, including traditional Convolutional Neural Networks (CNNs) and Multi-Layer Perceptrons (MLPs), map finite-dimensional vectors to finite-dimensional vectors. Consequently, they are inherently tied to the precise discretization of the data they are trained on. The advent of the Deep Operator Network (DeepONet)~\cite{lu2021learning} and the Fourier Neural Operator (FNO)~\cite{li2021fourier} pioneered the critical shift toward learning mappings between infinite-dimensional function spaces~\cite{roy2025pimrno}. By parameterizing the model in continuous function space, neural operators theoretically permit evaluation at any arbitrary spatial resolution without the need for computationally prohibitive retraining~\cite{li2021fourier}. 

However, while architectures like DeepONet are highly flexible in the output domain and allow the evaluation of the solution field at any continuous location within the desired domain, they historically impose strict constraints on the input space. Specifically, traditional DeepONets require all input functions to be discretized at identical, fixed sensor locations, fundamentally limiting practical applications where parameter fields are captured at varying fidelity~\cite{roy2025pimrno}. Similarly, standard FNOs, while achieving state-of-the-art accuracy and efficiency via the Fast Fourier Transform (FFT), require dense, uniform meshes to perform global linear transforms in the functional space~\cite{li2024multi}.

To overcome the lack of input-resolution independence, recent literature introduced the Resolution Independent Neural Operator (RINO)~\cite{roy2025pimrno}. RINO utilizes adaptive dictionary learning algorithms to deduce a set of continuous basis functions, parameterized as Implicit Neural Representations (INRs) or Sinusoidal Representation Networks (SIRENs). These continuous basis functions project arbitrarily sampled input signals, treated as point clouds, onto a finite-dimensional embedding space. This mapping explicitly decouples the input resolution from the architecture, achieving full resolution independence~\cite{bahmani2025resolution}. Similarly, Neural Green's Functions~\cite{yoo2026neural} have been proposed to extract point-wise features from volumetric point clouds, enabling generalization across irregular domain geometries by imitating the integral behavior of classical Green's functions. While RINO and Neural Green's Functions successfully decouple the input resolution, they frequently require complex auxiliary dictionary learning phases, substantial pre-computation of eigen-decompositions, or heavy multi-fidelity active learning protocols~\cite{li2024multi}.

\subsection{Physics-Informed and Geometry-Adaptive Surrogates}

The integration of physical laws into neural network architectures has revolutionized the simulation of complex systems. Physics-Informed Neural Networks (PINNs) integrate known physical principles into the optimization landscape by computing PDE residuals using automatic differentiation, effectively penalizing deviations from the governing equations~\cite{du2026aircraft}. However, scaling PINNs to highly irregular, parameterized geometric domains remains a profound challenge, as traditional PINNs often fail to produce meaningful solutions in complex shapes without extensive boundary sampling~\cite{costabal2024delta}.

To address the limitations of standard PINNs on complex geometries, PhyGeoNet~\cite{gao2021phygeonet} was introduced as a physics-informed, geometry-adaptive convolutional neural network capable of solving parameterized steady-state PDEs on irregular domains. PhyGeoNet achieves this by establishing a mapping between irregular physical domains and regular reference grids, allowing standard convolutional operators to function under physical constraints without requiring labeled data. While highly effective for specific boundary value problems, PhyGeoNet fundamentally relies on the construction of these reference grids, limiting its applicability to purely mesh-free or completely unstructured point-cloud paradigms. 

Further extending geometry-aware capabilities, Geometry-Informed Neural Operators (GINO) utilize signed distance functions (SDF) and point-cloud representations to handle highly varying 3D shapes, such as aerodynamic vehicle geometries~\cite{li2023geometry}. GINO transforms irregular point clouds into regular latent grids via graph neural operators, subsequently applying Fourier operations before decoding back to the physical space. This hybrid approach demonstrates discretization-convergence but retains the intermediate gridding step. Similarly, Physics-Informed Geometry-Aware Neural Operators (PI-GANO) and Physics-Informed Deep Compositional Operator Networks (PI-DCON) facilitate PDE inference under variable domain geometries by minimizing physical residuals over continuous fields without finite element (FEM) computation~\cite{zhong2025physics}. 

More recently, physics-informed multi-resolution neural operators have been formulated to enforce discrete PDEs through finite-difference constraints on arbitrary spatial resolutions. In these multi-resolution architectures, the operator takes a latent code along with spatiotemporal coordinates to produce continuous solutions in the physical space~\cite{roy2025pimrno}. While these methods show promise, the approach detailed herein expands on this trajectory by strictly maintaining a resolution-free coordinate encapsulation on grids of different scales from the initial geometric data encoding through the entire decoding process with interpolation to align features at different scales.

\subsection{Learning-Based Geometric Mapping and Variational Methods}

The computation of mappings that satisfy specific geometric distortions, such as density equalization or quasi-conformality, has historically relied on numerical diffusion solvers or explicit energy minimization techniques~\cite{gastner2004diffusion,Levy2002lscm}. Prior Density-Equalizing Maps (DEM) utilize Fick's law of diffusion to iteratively deform a mesh such that varying regional densities are uniformized, achieving area-preserving or density-prescribed parameterizations~\cite{choi2018density,lyu2024spherical}. Applications of DEM range from sociological data visualization (cartograms) to medical imaging and surface registration~\cite{lyu2024bijective}. However, conventional DEM techniques face several challenges: they may suffer from limited accuracy, produce overlapping artifacts in extreme density gradients, and require substantial algorithmic redesign when extending from 2D to 3D due to the derivative-dependent nature of their classical energy formulations~\cite{huang2025ldem}.

To bridge density equalization and quasi-conformality, Lyu et al.~\cite{lyu2024bijective} developed the Density-Equalizing Quasi-conformal (DEQ) mapping method, achieving a balance between strict density equalization and angle-preserving quasi-conformality for simply and multiply connected surfaces. More recently, Learning-based Density-Equalizing Mapping (LDEM) frameworks~\cite{huang2025ldem} have been proposed that utilize deep neural networks to directly minimize a loss function enforcing density uniformity and geometric regularity. LDEM utilizes a hierarchical approach to predict transformations at coarse and dense levels, easily generalizing from 2D to 3D domains without altering structural architectures. 

Similarly, quasi-conformal mappings, which bound angular distortion via the Beltrami equation, have seen a transition toward deep learning. Traditional approaches minimize the norm of the Beltrami coefficient (the Beltrami energy) to achieve optimal quasi-conformal parameterizations (QCMC) of multiply-connected domains~\cite{ho2016qcmc}. Deep learning frameworks based on Teichm\"uller theory now learn the Beltrami coefficient directly as a latent feature vector, yielding diffeomorphic mappings in real-time by integrating Beltrami-based regularization into the network's loss function~\cite{chen2024deep}. Our framework encodes samples of the parameter field and their spatial coordinates as channels on a structured grid. It combines features from multiple grid scales and reuses the same trained weights at different supported grid sizes, which could evaluate geometry-aware objectives on structured grids of several sizes. Table~\ref{tab:overview} summarizes the input representations and objectives.

\begin{table}[t]
\centering
\resizebox{\columnwidth}{!}{
\begin{tabular}{l l l l}
\hline
\textbf{Framework} & \textbf{Core Architecture} & \textbf{Input Resolution} & \textbf{Constraint} \\
\hline
FNO~\cite{li2021fourier} & Global Fourier Transforms & Uniform grid & Data-driven / PINN \\
DeepONet~\cite{lu2021learning} & Branch/Trunk MLPs & Fixed sensor locations & Data-driven / PINN \\
RINO~\cite{bahmani2025resolution} & INR/SIREN projection & Dictionary embedding & Data-driven mappings \\
PhyGeoNet~\cite{gao2021phygeonet} & Hierarchical CNN & Regular reference grids & Physics-informed \\
GINO~\cite{li2023geometry}& Graph + Fourier layers & Point cloud to latent grid & SDF-based \\
LDEM~\cite{huang2025ldem} & Hierarchical CNN & Fixed-scale mesh & Density loss \\
\textbf{Proposed} & Modified U-Net & Structured grid, variable size & Variational energy \\
\hline
\end{tabular}
}
\caption{Overview and comparison of different operator learning frameworks.}
\label{tab:overview}
\end{table}

\section{Mathematical Background}
\label{sect:background}

To formalize the proposed data-free neural surrogate, we first establish the mathematical foundations of the governing physical and geometric constraints. Specifically, we review surface parameterization and distortion measures, quasi-conformal theory based on the Beltrami equation, and diffusion-driven density-equalizing mappings.

\subsection{Quasi-Conformal Theory and the Beltrami Equation}

An orientation-preserving quasi-conformal homeomorphism $f$ satisfies the Beltrami equation:
\begin{equation}
f_{\bar{z}} = \mu(z) f_z,
\label{eq:beltrami_equation}
\end{equation}
where $\mu(z)$ is the Beltrami coefficient. Here, the Wirtinger derivatives are defined as $f_z = \frac{1}{2}(f_x - i f_y)$ and $f_{\bar{z}} = \frac{1}{2}(f_x + i f_y)$.
Thus, we have $\mu(z) = \frac{f_{\bar{z}}}{f_z}$.

\begin{figure}[t]
    \centering
    \includegraphics[width=0.8\linewidth]{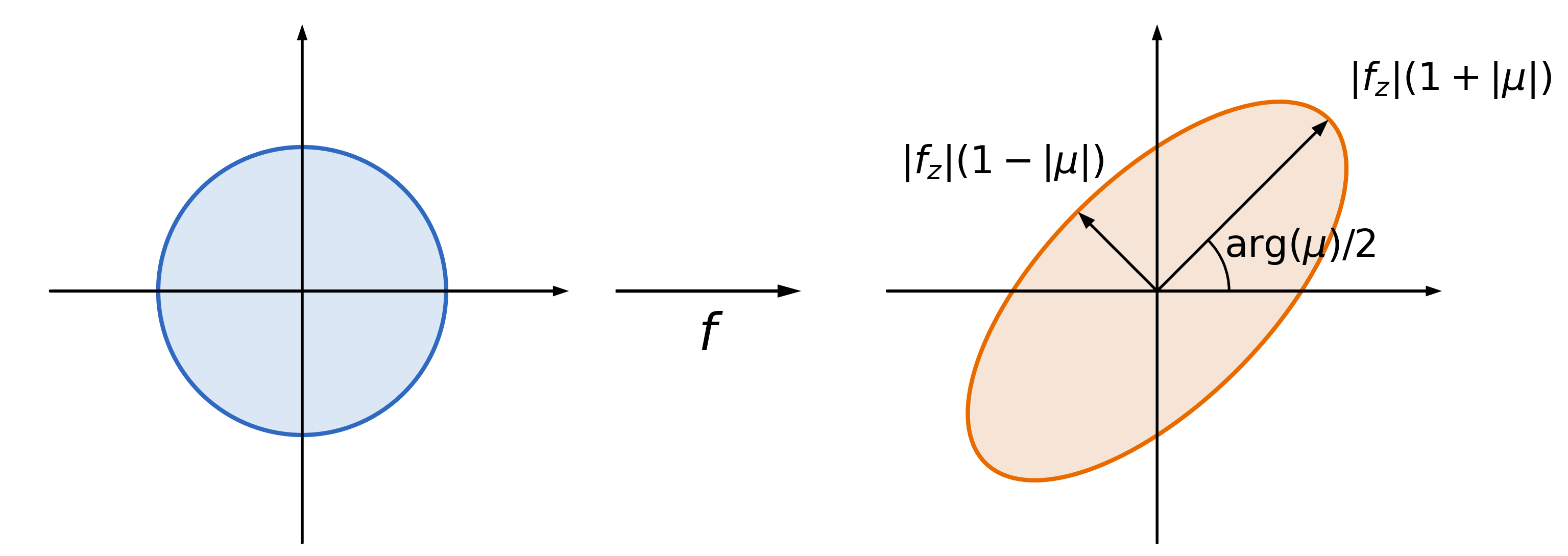}
    \caption{
    \textbf{Illustration of the quasi-conformal distortion induced by the Beltrami coefficient $\mu$ for a map $f$.} 
    The image of an infinitesimal circle under $f$ is an infinitesimal ellipse with maximal and minimal stretch factors $|f_z|(1+|\mu|)$ and $|f_z|(1-|\mu|)$. For $\mu\ne0$, the direction of maximal stretch in the source coordinates has angle $\arg(\mu)/2$.
    }
    \label{fig:bc_sample}
\end{figure}

A conformal map has $\mu = 0$. At a point where $f$ is differentiable and $f_z\ne 0$, the Jacobian determinant is $\det(J_f)=|f_z|^2(1-|\mu_f|^2)$. It is positive when $|\mu_f|<1$, indicating local orientation preservation. The geometric distortion of a quasi-conformal map $f$ is quantified by the pointwise dilatation (see Fig.~\ref{fig:bc_sample} for an illustration):
\begin{equation}
K_f(z) = \frac{1 + |\mu_f(z)|}{1 - |\mu_f(z)|}.
\label{eq:maximal_dilatation}
\end{equation}

On a bounded Jordan domain, Theorem~1 of Chen et al.~\cite{chen2026circular} gives a global criterion. A map continuous on the closure and $C^1$ in the interior, with $f_z\ne0$ and $\|\mu_f\|_{L^\infty}<1$, is a homeomorphism onto the closed region enclosed by its boundary image if the boundary restriction is an orientation-preserving homeomorphism onto a Jordan curve. The interior conditions give $\det(J_f)>0$, and the boundary condition fixes the global correspondence.

The Beltrami energy is defined as:
\begin{equation}
\mathcal{E}_B(f) = \int_{\Omega} |\mu_f(z)|^2 \, dz.
\label{eq:beltrami_energy}
\end{equation}
Minimizing this energy yields mappings that are as conformal as possible under given constraints.

For quasi-conformal homeomorphisms $f$ and $g$ on a common source domain, the Beltrami coefficient satisfies:
\begin{equation}
\mu(f \circ g^{-1}) \circ g =
\frac{\mu(f) - \mu(g)}{1 - \mu(f)\overline{\mu(g)}} \cdot \frac{g_z}{\overline{g_z}}.
\end{equation}

These formulations form the theoretical foundation for enforcing geometric constraints in neural surrogate training.

\subsection{Surface Parameterization and Mapping Distortion Measures}

A surface parameterization is a one-to-one mapping $u : \Omega \rightarrow S$ from a parameter domain $\Omega \subset \mathbb{R}^2$ to a surface $S$. For the mapping $u(u_1,u_2)$ to be regular, the tangent vectors $x_1 = \frac{\partial u}{\partial u_1}$, $x_2 = \frac{\partial u}{\partial u_2}$ must remain linearly independent at every point on the surface.

The local geometric properties of the parameterization are described by the first fundamental form, or metric tensor, $g_{\alpha\beta} = x_\alpha \cdot x_\beta$, which determines the infinitesimal arc-length element: $ds^2 = \sum_{\alpha,\beta} g_{\alpha\beta}\, du_\alpha du_\beta$.

Distortion introduced by a mapping is characterized by deviations of the metric tensor from the Euclidean metric. An isometric mapping preserves lengths, angles, and areas, requiring $g = I$, where $I$ is the identity matrix. Since exact isometric flattening is generally only possible for developable surfaces, practical parameterization methods instead aim to preserve selected geometric properties, such as angles or areas.

Conformal mappings preserve local angles, allowing only isotropic scaling. Their metric tensor takes the form $g = \eta I$, where $\eta$ is a scalar scaling factor. In contrast, equiareal, or authalic, mappings preserve surface area by maintaining the area element $dA = \sqrt{\det g}\, du_1 du_2$. Such mappings are particularly important in applications requiring density preservation or area-based analysis.

In this work, mapping distortion is evaluated using two complementary quantities: the Beltrami coefficient and the Jacobian determinant. The magnitude of the Beltrami coefficient, $|\mu_f|$, is used to measure local angular distortion, where smaller values indicate that the mapping is closer to conformal. The corresponding pointwise dilatation and global Beltrami energy are previously introduced in Eq.~\eqref{eq:maximal_dilatation} and Eq.~\eqref{eq:beltrami_energy}, respectively.

The Jacobian determinant is used to measure local area change. For a planar mapping $f : \Omega \rightarrow \Omega'$ written as $f(x,y) = (f_1(x,y), f_2(x,y))$, the Jacobian matrix is
\begin{equation}
J_f =
\begin{pmatrix}
\frac{\partial f_1}{\partial x} & \frac{\partial f_1}{\partial y} \\
\frac{\partial f_2}{\partial x} & \frac{\partial f_2}{\partial y}
\end{pmatrix},
\end{equation}
and the signed local area scale is given by $\det(J_f)$. When $\det(J_f)=1$, the mapping locally preserves area. Values larger than one indicate local expansion, while values between zero and one indicate local shrinkage. A positive Jacobian determinant indicates local orientation preservation. For a piecewise-affine map, strictly positive signed determinants on every simplex and a bijective boundary correspondence give global bijectivity~\cite[Theorem~1]{lipman2014bijective}. Centered differences at grid nodes measure a different quantity from these elementwise determinants.

For surface parameterizations, the area scaling is closely related to the determinant of the metric tensor. The quantity $\sqrt{\det g}$ represents the local surface area element induced by the parameterization. Therefore, deviations of $\sqrt{\det g}$, or equivalently the Jacobian determinant in the planar representation, provide a natural measurement of area distortion.

Together, the Beltrami coefficient and the Jacobian determinant capture two different aspects of mapping distortion. The Beltrami coefficient reflects angular or conformal distortion, while the Jacobian determinant reflects area distortion and local density change. These two quantities are therefore used as the main distortion measures in our mapping framework, especially when both geometric regularity and density preservation are required.

\subsection{Diffusion Equations and Density-Equalizing Maps}

\begin{figure}[t]
    \centering
    \includegraphics[width=0.8\linewidth]{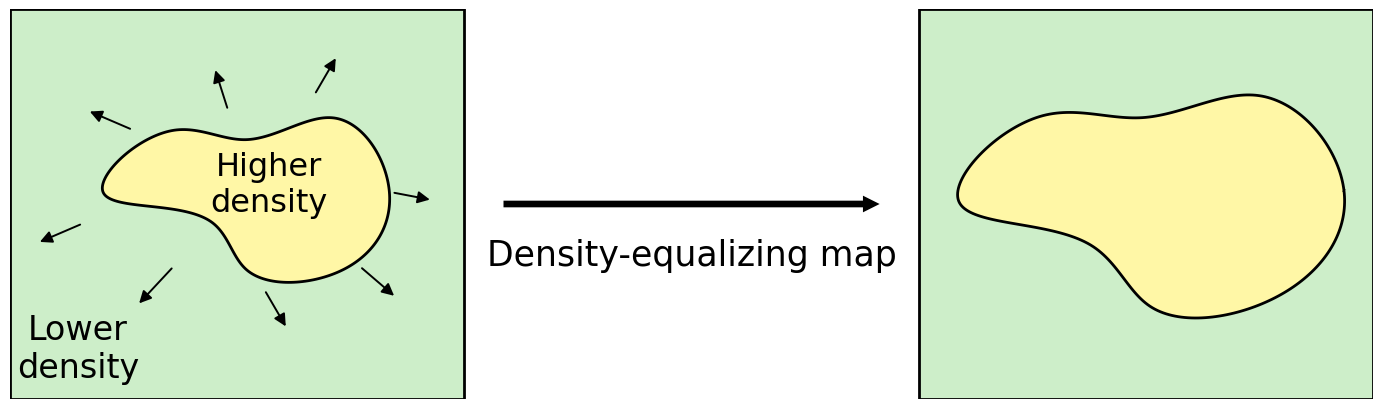}
    \caption{
    \textbf{An illustration of the density-equalizing map.} Under a density-equalizing map, regions with higher density expand outward into surrounding lower-density areas until the density becomes more uniformly distributed across the mapped domain.
    }
    \label{fig:dem_example}
\end{figure}

Density-equalizing maps are based on diffusion processes governed by:
\begin{equation}
\frac{\partial \rho}{\partial t}(\mathbf{x}, t) = D \, \Delta \rho(\mathbf{x}, t),
\end{equation}
where $\rho$ is the density field and $D$ is the diffusion coefficient. A no-flux boundary condition ensures mass conservation: $\nabla \rho \cdot \mathbf{n} = 0$ on $\partial \Omega$.
Without loss of generality, the diffusion coefficient can be absorbed through a rescaling of the time variable ($t \leftarrow Dt$), effectively setting $D = 1$. Under this normalized time scale, the deformation is driven by a velocity field derived from Fick's law:
\begin{equation}
\mathbf{v}(\mathbf{x}, t) = -\frac{\nabla \rho(\mathbf{x}, t)}{\rho(\mathbf{x}, t)}.
\end{equation}
The mapping evolves as:
\begin{equation}
\mathbf{x}(t) = \mathbf{x}(0) + \int_0^t \mathbf{v}(\mathbf{x}(\tau), \tau)\, d\tau.
\end{equation}
At steady state, the density converges to a uniform distribution (see Fig.~\ref{fig:dem_example}):
\begin{equation}
\rho(\mathbf{x}, t) \to \bar{\rho} = \frac{1}{|\Omega|} \int_\Omega \rho(\mathbf{x}, 0)\, d\mathbf{x}.
\end{equation}

\section{Proposed Method}
\label{sect:method}

\subsection{Geometric Data Encoding}
\label{subsec:data_encoding}

To enable resolution-free learning of geometric parameterization and mapping problems, the neural surrogate operates on a unified tensor representation that jointly encodes spatial coordinates and problem-dependent geometric fields. Rather than relying on fixed discretizations or handcrafted grid-dependent features, the proposed framework samples this field on a structured grid and stacks the samples as a $(d+m)$-channel tensor. The resulting representation supplies both the spatial coordinates and the parameter values to the network.

Let $\Omega \subset \mathbb{R}^d$ denote the computational domain, where $d \ge 2$ is the spatial dimension, and let $\mathbf{x} = (x_1,\ldots,x_d) \in \Omega$ denote spatial coordinates. We consider a collection of scalar or vector-valued parameter fields $\boldsymbol{p}(\mathbf{x}) = \bigl(p_1(\mathbf{x}), \ldots, p_m(\mathbf{x})\bigr)$, which may represent densities, material parameters, source terms, or other problem-specific quantities. The input to the neural surrogate is constructed as a stacked multi-channel field
\begin{equation}
\mathcal{U}(\mathbf{x})
=
\bigl[
x_1,\ldots,x_d,\,
p_1(\mathbf{x}),\ldots,p_m(\mathbf{x})
\bigr] = [\mathbf{x},\boldsymbol{p}(\mathbf{x})],
\end{equation}
which is subsequently sampled on a grid of arbitrary resolution and treated as a $(d+m)$-channel tensor. This encoding allows the network to access both geometric location information and problem parameters in a uniform manner, analogous to a multi-channel image.

To ensure broad coverage of geometric configurations during training, the parameter fields $\boldsymbol{p}$ are generated as continuous random fields on $\Omega$. A typical construction combines smooth multi-frequency components with localized non-smooth perturbations. The smooth component may be represented as a finite expansion
\begin{equation}
p_{\mathrm{smooth}}(\mathbf{x})
=
\sum_{k=1}^{K}
A_k
\prod_{i=1}^{d}
\sin\left(2\pi f_{k,i} x_i + \varphi_{k,i}\right),
\end{equation}
where amplitudes $A_k$, frequencies $f_{k,i}$, and phases $\varphi_{k,i}$ are randomly sampled over prescribed ranges. Frequencies are often drawn from log-uniform distributions to capture both low- and high-frequency spatial variations.

To mimic realistic geometric irregularities, non-smooth features are introduced locally. These may include additive noise, discontinuous step regions, or compact high-frequency patches supported on small subsets of the domain. The resulting parameter field can be written as
\begin{equation}
p(\mathbf{x})
=
p_{\mathrm{smooth}}(\mathbf{x})
+
p_{\mathrm{nonsmooth}}(\mathbf{x}),
\end{equation}
where $p_{\mathrm{nonsmooth}}$ is sparse in space and varies randomly across samples. This construction exposes the network to both smooth regions and sharp geometric features, which frequently arise in practical geometric PDEs.

For each training instance, the continuous field $\mathcal{U}(\mathbf{x})$ is sampled on a grid $\{\mathbf{x}_{\boldsymbol{i}}\}_{\boldsymbol{i} \in \mathcal{I}_h}
\subset \Omega$, where the grid resolution is randomly selected within a prescribed range. This randomized sampling strategy exposes the network to more than one grid size during training. When required by the physical model, parameter fields may be normalized using discrete quadrature weights. For example, on a uniform grid with spacing $h$,
\[
\int_\Omega p(\mathbf{x})\,\mathrm{d}\mathbf{x}
\;\approx\;
\sum_{\boldsymbol{i}} p(\mathbf{x}_{\boldsymbol{i}})\, h^d,
\]
The normalization used in each experiment is specified in Section~\ref{sect:experiment}.

The proposed encoding strategy naturally extends to higher-dimensional and multi-parameter settings. Since spatial coordinates and parameter fields are treated uniformly as channels, the same network architecture and training pipeline apply across dimensions and problem classes.

This geometric data encoding framework provides a flexible and resolution-independent interface between continuous geometric problems and convolutional neural surrogates. By operating directly on coordinate-augmented fields, the model learns mappings that generalize across spatial discretizations, dimensions, and parameter configurations.

\subsection{Model Configuration}
\label{subsect:model_config}

\begin{figure}[t]
    \centering
    \includegraphics[width=1.0\linewidth]{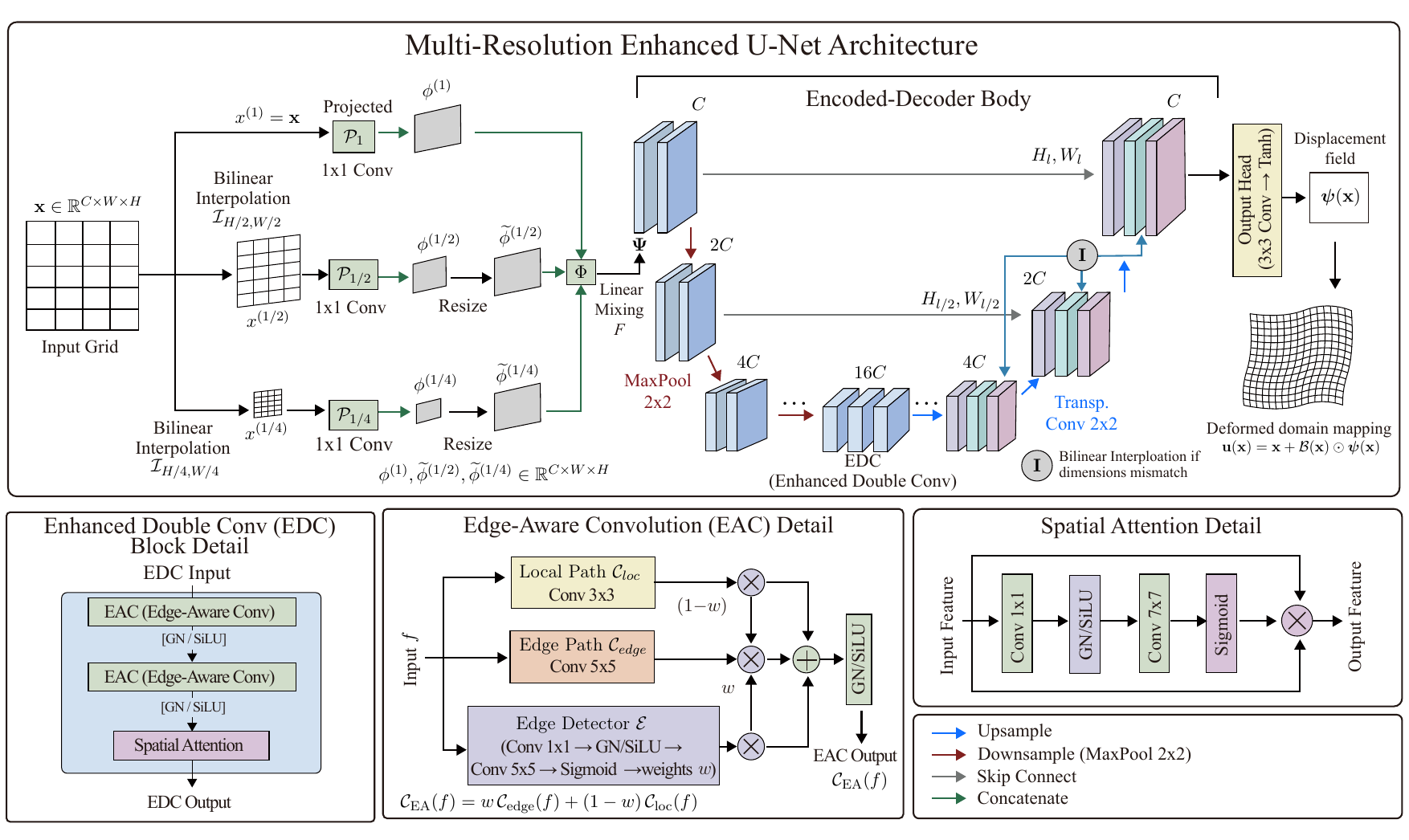}
    \caption{\textbf{Architectural diagram of the Multi-Resolution Enhanced U-Net.} The top panel illustrates the overall Encoder-Decoder body with multi-scale input resolution processing. The bottom panels provide detailed breakdowns of the specialized sub-modules: the Enhanced Double Conv (EDC) block, the Edge-Aware Convolution (EAC) detail and the Spatial Attention mechanism.}
    \label{fig:modelstructure}
\end{figure}

We construct a resolution-free neural surrogate based on an enhanced U-Net architecture (Fig.~\ref{fig:modelstructure}), whose design allows the same trained weights to be used at different supported grid sizes. The model integrates multi-resolution input processing, interpolation-based resolution alignment, and edge-aware convolutional operators to approximate solution mappings of geometric partial differential equations and variational problems.

Let $\mathbf{x} \in \mathbb{R}^{C \times H \times W}$ denote the discretized input field on a structured grid of size $(H,W)$. To capture both global structure and fine-scale variations, the model processes the input at multiple spatial scales. In addition to the original resolution $x^{(1)} = \mathbf{x}$, we construct two coarser representations via interpolation,
\[
x^{(1/2)} = \mathcal{I}_{H/2,W/2}(\mathbf{x}),
\qquad
x^{(1/4)} = \mathcal{I}_{H/4,W/4}(\mathbf{x}),
\]
where $\mathcal{I}_{h,w}(\cdot)$ denotes bilinear interpolation to resolution $(h,w)$.

Each resolution is independently projected into a feature space through a pointwise linear operator,
\begin{equation}
\phi^{(1)} = \mathcal{P}_1\!\left(x^{(1)}\right), \qquad
\phi^{(1/2)} = \mathcal{P}_{1/2}\!\left(x^{(1/2)}\right), \qquad
\phi^{(1/4)} = \mathcal{P}_{1/4}\!\left(x^{(1/4)}\right),
\end{equation}
where $\mathcal{P}_\ast : \mathbb{R}^{C} \rightarrow \mathbb{R}^{C_\ast}$ are pointwise linear mappings realized as $1\times1$ convolutions. These operators act independently at each spatial location and therefore introduce no coupling across grid points. Their purpose is to lift input fields at different resolutions into compatible feature spaces while preserving spatial locality.

Since $\phi^{(1/2)}$ and $\phi^{(1/4)}$ are defined on coarser grids, they are interpolated back to the original resolution $(H,W)$,
\begin{equation}
\widetilde{\phi}^{(1/2)} = \mathcal{I}_{H,W}\!\left(\phi^{(1/2)}\right), \qquad
\widetilde{\phi}^{(1/4)} = \mathcal{I}_{H,W}\!\left(\phi^{(1/4)}\right).
\end{equation}
The three feature fields are then concatenated along the channel dimension to form a composite representation
\begin{equation}
\Phi =
\bigl[
\phi^{(1)},\,
\widetilde{\phi}^{(1/2)},\,
\widetilde{\phi}^{(1/4)}
\bigr]
\in \mathbb{R}^{C_{\mathrm{tot}} \times H \times W},
\end{equation}
where $C_{\mathrm{tot}} = C_1 + C_{1/2} + C_{1/4}$. This construction aggregates information from full, half, and quarter resolutions at each spatial location, enabling local features to incorporate global geometric context encoded at coarser scales.

A linear mixing operator $\mathcal{F}$ is subsequently applied to $\Phi$ to produce a unified embedding,
\begin{equation}
\Psi = \mathcal{F}(\Phi),
\end{equation}
where $\mathcal{F}$ is implemented as a $1\times1$ convolution. At each spatial location $(i,j)$, this operation takes the form
\[
\Psi(:,i,j) = W\,\Phi(:,i,j),
\]
with $W$ a trainable weight matrix shared across all grid points. This pointwise linear mixing preserves spatial locality while adaptively fusing information across resolutions. The same interpolation and pointwise linear operations apply at different grid sizes, while the resulting embedding depends on the grid samples.

The fused feature field $\Psi$ is then processed by an encoder–decoder architecture with skip connections. In the encoder, spatial resolution is progressively reduced using pooling operators, while the channel dimension is increased to encode higher-level geometric information. Each encoding stage applies enhanced convolutional blocks designed to distinguish smooth interior regions from sharp interfaces and geometrically complex boundaries.

In the decoder, feature maps are upsampled using transposed convolution operators that nominally increase spatial resolution by a factor of two at each stage. Since the input resolution $(H,W)$ is arbitrary, the spatial dimensions produced by transposed convolutions may not exactly match those of the corresponding encoder features. To ensure consistent alignment, bilinear interpolation is applied whenever necessary so that the upsampled feature map satisfies $\Psi_{\mathrm{up}} \in \mathbb{R}^{\ast \times H_\ell \times W_\ell}$, where $(H_\ell,W_\ell)$ coincides with the resolution of the associated encoder feature. The aligned features are then concatenated through skip connections and refined by enhanced convolutional blocks, restoring fine-scale geometric details attenuated during downsampling.

A central component of each convolutional block is an edge-aware convolution operator. Given an intermediate feature field $f$, a learnable spatial weighting function $w = \sigma\!\bigl(\mathcal{E}(f)\bigr)$
is computed, where $\mathcal{E}$ denotes a local edge-detection operator and $\sigma$ is the sigmoid function. Two convolutional responses are evaluated: a local operator $\mathcal{C}_{\mathrm{loc}}$ and an edge operator $\mathcal{C}_{\mathrm{edge}}$, labeled with $3\times3$ and $5\times5$ kernels, respectively, in Fig.~\ref{fig:modelstructure}. These responses are adaptively combined as
\begin{equation}
\mathcal{C}_{\mathrm{EA}}(f)
=
w \, \mathcal{C}_{\mathrm{edge}}(f)
+
(1-w)\,\mathcal{C}_{\mathrm{loc}}(f),
\end{equation}
allowing the network to adjust its effective receptive behavior according to local geometric complexity.

Finally, the decoder produces a vector-valued displacement field $\boldsymbol{\psi}(\mathbf{x}) = \mathcal{N}_\theta(\mathbf{x})$ defined on the same spatial resolution as the input grid, where $\mathcal{N}_\theta$ denotes the neural surrogate operator and $\theta$ is the corresponding training parameters. The mapped domain is obtained through an explicit deformation of the reference coordinates,
\begin{equation}
\mathbf{u}(\mathbf{x})
=
\mathbf{x}
+
\mathcal{B}(\mathbf{x}) \odot \boldsymbol{\psi}(\mathbf{x}),
\end{equation}
where $\mathcal{B}(\mathbf{x})$ is a boundary mask enforcing prescribed constraints and $\odot$ denotes component-wise multiplication. The choice of the mask depends on the prescribed boundary conditions. For example, on the unit square, $\mathcal{B}(\mathbf{x})=\mathbf{x}\odot(\mathbf1-\mathbf{x})$ fixes the normal coordinate on each side and both coordinates at the four corners, while allowing tangential motion along the edges. For free-boundary cases, we use $\mathcal{B}(\mathbf{x})=\mathbf{1}$, which leaves all displacement components unconstrained by the mask. Other masks can be chosen to impose different componentwise displacement constraints.

Let $X_h$ denote the structured input grid and $p_h$ the sampled parameter field. The learned operator maps $p_h$ to the sampled deformation $u_h$ conditional on $X_h$. Interpolation aligns the internal feature maps so that the network returns one displacement per input grid node. The implementation takes structured tensors as input and returns predictions at those nodes. Direct evaluation from scattered samples or at off-grid query points falls outside this implementation. The physical extent of the convolution stencil and the discrete objective can vary with the grid spacing. We assess accuracy at the tested grid sizes in Section~\ref{sect:experiment}.

\subsection{Weight Refinement}
\label{subsect:weight_refinement}

Although the proposed neural surrogate provides resolution-free approximations of geometric solution fields, certain geometric partial differential equations and variational problems are sensitive to anisotropic stretching effects along different coordinate directions. To improve solution accuracy and preserve the boundary conditions without modifying network parameters, we introduce a post-processing weight refinement procedure that optimizes a low-dimensional vector of stretching parameters applied to the network output.

Let $\mathbf{x} \in \mathbb{R}^d$ denote spatial coordinates in dimension $2 \leq d \leq 3$, and let $\boldsymbol{\phi}(\mathbf{x}) = \bigl(\phi_1(\mathbf{x}),\ldots,\phi_d(\mathbf{x})\bigr)$ be the displacement field predicted by the neural surrogate model. We consider a family of refined mappings parameterized by a stretching vector $\boldsymbol{\alpha}
= (\alpha_1,\ldots,\alpha_d) \in \mathbb{R}^d$, defined as
\begin{equation}
\mathbf{u}_{\boldsymbol{\alpha}}(\mathbf{x})
=
\mathbf{x}
+
\mathcal{B}(\mathbf{x})
\odot
\bigl(\boldsymbol{\alpha} \odot \boldsymbol{\phi}(\mathbf{x})\bigr),
\end{equation}
where $\odot$ denotes component-wise multiplication and $\mathcal{B}(\mathbf{x})$ is a spatial mask enforcing prescribed boundary conditions. Each component $\alpha_i$ independently controls the stretching magnitude along the $i$-th coordinate direction, allowing anisotropic adjustment of the deformation field.

The optimal stretching vector is obtained by minimizing a problem-dependent objective functional,
\begin{equation}
\boldsymbol{\alpha}^\ast
=
\arg\min_{\boldsymbol{\alpha} \in \mathcal{A}}
\;
\mathcal{J}\!\left(\mathbf{u}_{\boldsymbol{\alpha}}\right),
\end{equation}
where $\mathcal{A} \subset \mathbb{R}^d$ is the searching domain and $\mathcal{J}$ measures deviation from the governing geometric constraint or variational principle. Depending on the application, $\mathcal{J}$ may encode residuals of geometric invariants, energy discrepancies, or constraint violations.

To ensure admissibility of the refined mapping, the objective may incorporate regularization terms penalizing undesirable geometric behavior,
\begin{equation}
\mathcal{J}\!\left(\mathbf{u}\right)
=
\mathcal{L}_{\mathrm{geom}}\!\left(\mathbf{u}\right)
+
\lambda \, \mathcal{P}\!\left(\mathbf{u}\right),
\end{equation}
where $\mathcal{L}_{\mathrm{geom}}$ measures the task discrepancy and $\mathcal{P}$ penalizes selected local violations. The raw prediction corresponds to $\boldsymbol{\alpha}=\mathbf{1}$. The reference identity corresponds to $\boldsymbol{\alpha}=\mathbf{0}$.

Since the dimension $d$ is small (typically $d=2$ or $3$), the optimization over $\boldsymbol{\alpha}$ remains low-dimensional and computationally inexpensive. In practice, derivative-free line-search strategies can be employed, such as alternating one-dimensional searches over each component $\alpha_i$ or direct multi-dimensional variants of golden-section-type methods. These approaches require only evaluations of $\mathcal{J}$ and are robust to nonconvexity and nondifferentiability of the objective.

The refined mapping $\mathbf{u}_{\boldsymbol{\alpha}^\ast}$ preserves the resolution-free nature of the neural surrogate and naturally extends to both planar and volumetric settings. By allowing independent directional stretching, the proposed refinement provides a flexible and efficient mechanism for improving geometric consistency in high-dimensional problems without additional network training.

\section{Experimental Results} 
\label{sect:experiment}

\subsection{Reconstruction of Mapping from Beltrami Coefficient}
Density-equalization problems and quasi-conformal mappings are intimately connected to the \textbf{Beltrami equation}, a fundamental geometric PDE that governs the distortion of mappings between domains. Given a complex dilatation (or Beltrami coefficient) $\mu(z)$ defined on a planar domain $\Omega$, the Beltrami equation seeks a homeomorphism $f:\Omega \to \Omega$ satisfying Eq.~\eqref{eq:beltrami_equation} where $\mu(z)$ encodes local conformal distortion. Solutions to this equation play a central role in conformal geometry and applications such as surface parameterization and medical imaging. Here, we synthesize randomized Beltrami coefficients $\mu(z)$ to train a neural surrogate for solving the Beltrami equation under boundary constraints.

At each training iteration, we first generate a map $G$ on a uniform grid of the unit square $\Omega=[0,1]^2$ and then compute its Beltrami coefficient. The smooth part of a random displacement field is
\begin{equation*}
T(x,y)
=\sum_{i=1}^{n} A_i \sin\!\left(2\pi f_{x,i}x+\phi_{x,i}\right)
                 \cos\!\left(2\pi f_{y,i}y+\phi_{y,i}\right).
\end{equation*}
At each of the $K$ deformation substeps used to construct a single training input, we independently draw one horizontal and one vertical displacement field of this form. We sample $n$ uniformly from the integers $5$ through $35$, amplitudes $A_i$ uniformly from $[-1,1]$, frequencies $f_{x,i},f_{y,i}$ log-uniformly from $[0.1,12]$, and phases $\phi_{x,i},\phi_{y,i}$ uniformly from $[0,2\pi]$. We may add perturbations before dividing each field by its sample standard deviation plus $10^{-6}$. We may then add shear or rotation to the pair. The resulting fields at substep $j$ are denoted by $T_{j,1}$ and $T_{j,2}$. We sample $K$ uniformly from the integers $2$ through $6$. Starting from the identity, we construct the map at interior grid nodes by
\begin{equation*}
\begin{aligned}
G^{(j)}(\mathbf{x})
&=\operatorname{clip}_{[0,1]^2}\!\left(
G^{(j-1)}(\mathbf{x})+
\eta\,\mathcal{B}(\mathbf{x})\odot (T_{j,1},T_{j,2})\!\left(G^{(j-1)}(\mathbf{x})\right)
\right),\\
G^{(0)}(\mathbf{x})&=\mathbf{x},\qquad j=1,\ldots,K,\qquad G=G^{(K)}.
\end{aligned}
\end{equation*}

The componentwise displacement mask $\mathcal{B}(\mathbf{x}) = \mathbf{x}\odot(\mathbf1-\mathbf{x})$ fixes the corners and the normal coordinate on each side of the square, allowing tangential motion along the edges. Fixing the target square and corner correspondences removes the freedom to postcompose a quasi-conformal homeomorphism with a conformal map. A strictly monotone correspondence along each edge gives the boundary homeomorphism required by the planar bijectivity criterion.

We use bilinear interpolation to evaluate the displacement fields at the current mapped position. Clipping keeps each coordinate in $[0,1]$, and boundary nodes are reset to their original positions after every update. We draw the initial step size $\eta$ uniformly from $[0.02,0.08]$ and use the same $\eta$ for all $K$ updates in a trial. We compute the Beltrami coefficient of the resulting map $G=(G_1,G_2)$ from $G_1+iG_2$ by centered differences at interior nodes,
\begin{equation}
\label{eq:train_beltrami}
\mu(x,y)=\mu_R(x,y)+i\,\mu_I(x,y)
=\frac{D_{\bar z,h}G(x,y)}{D_{z,h}G(x,y)+10^{-8}},
\end{equation}
where $D_{z,h}$ and $D_{\bar z,h}$ are discrete Wirtinger derivatives. If a trial fails the checks, we multiply the step size by $0.7$ and restart from the identity using the same displacement fields. We accept the map when the sampled interior coefficient satisfies $\max|\mu|\leq0.75$ and the minimum sampled Jacobian determinant exceeds $10^{-3}$. If none of the $25$ trials passes these checks, we use the identity. We fill the boundary entries of the coefficient array with zeros. The network receives the four channels $(x,y,\mu_R,\mu_I)$. The generated map $G$ serves only to construct the input coefficient.

The neural surrogate is trained using the following configuration:
\begin{itemize}
    \item \textbf{Epochs:} $250000$
    \item \textbf{Optimizer:} AdamW with initial learning rate $10^{-4}$ and weight decay $10^{-5}$
    \item \textbf{Learning-rate schedule:} Cosine annealing over $250000$ updates with minimum learning rate $10^{-6}$
    \item \textbf{Training grid resolution:} One uniform $N_{\mathrm{train}}\times N_{\mathrm{train}}$ grid per update, with $N_{\mathrm{train}}$ sampled uniformly from the integers $64$ through $96$
    \item \textbf{Network input:} the four-channel field $(x,\, y,\, \mu_R,\, \mu_I)$, so a complex coefficient is represented by two real channels
    \item \textbf{Gradient clipping:} $\ell^2$-norm clipped at 1.0
    \item \textbf{Hardware:} Single NVIDIA RTX~A6000 GPU; cuDNN disabled
\end{itemize}

During training, we optimize the network weights using a Beltrami-dependent anisotropic Dirichlet energy evaluated on the predicted mapping $(u,v)$, with regularization. During refinement, we compare the prescribed coefficient with the coefficient induced by the prediction using
\[
\mathcal L_{\mathrm{ref}}(u,v)=
\bigl\|\mu_{pred}(u,v)-\mu_{truth}\bigr\|_{2,h}^2+\mathcal R_{\mathrm{ref}}(u,v).
\]
Here $\|\cdot\|_{2,h}^2$ is the mean squared complex modulus over the interior grid nodes, and $\mathcal R_{\mathrm{ref}}$ collects the regularization terms.

For evaluation, the input coefficient of each test is the complex Beltrami coefficient induced by an analytically defined mapping $T_i$, computed using the same centered differences and boundary padding. The same four-channel representation $(x,y,\operatorname{Re}\mu_{T_i},\operatorname{Im}\mu_{T_i})$ is used. The mappings $T_i$ are held out for evaluation of the reconstructed displacement. The refinement objective is set to be
\[
\mathcal{J}=\|\mu_{\mathrm{pred}}(u,v)-\mu_{\mathrm{truth}}\|_{2,h}^{2}
+10^{-2}\max\{10^{-3}-\det J,\,0\}
+10\max\{|\mu_{\mathrm{pred}}|-0.98,\,0\}.
\]
The four transformations $T_i(x,y)=(u_i(x,y),v_i(x,y))$ are:

\begin{enumerate}[(a)]
    \item \textbf{Mild Sinusoidal Perturbation}: We have
    \begin{equation}
    \label{eq:test_map_T1}
    \begin{aligned}
    u_1(x,y) &= x + a\sin(\pi x)\sin(2\pi y), \ \  v_1(x,y) = y + b\sin(2\pi x)\sin(\pi y),
    \end{aligned}
    \end{equation}
    with $a=b=0.08$.

    \item \textbf{Multi-frequency Sinusoidal Deformation}: We have
    \begin{equation}
    \label{eq:test_map_T2}
    \begin{aligned}
    u_2(x,y) &= x
    + a\sin(2\pi y)\sin(\pi x)
    + b\sin(4\pi y)\sin(2\pi x),\\
    v_2(x,y) &= y
    + a\sin(2\pi x)\sin(\pi y)
    + b\sin(4\pi x)\sin(2\pi y),
    \end{aligned}
    \end{equation}
    with $a=0.03$ and $b=0.02$.

    \item \textbf{Localized Rotational Deformation}: We have
    \begin{equation}
    \label{eq:test_map_T3}
\resizebox{0.8\textwidth}{!}{$
    \begin{aligned}
    u_3(x,y) &= x + s(x,y)\left[
    -0.60\left(y-\tfrac12\right)e^{-r^2/0.045}
    +0.70\left(y-\tfrac12\right)e^{-r^2/0.045}
    +0.06\sin(2\pi y)
    \right],\\
    v_3(x,y) &= y + s(x,y)\left[
    0.60\left(x-\tfrac12\right)e^{-r^2/0.045}
    -0.70\left(x-\tfrac12\right)e^{-r^2/0.045}
    -0.06\sin(2\pi x)
    \right],
    \end{aligned}
    $}
    \end{equation}
    where $s(x,y)=\sin(\pi x)\sin(\pi y)$ and
    $r^2=(x-\tfrac12)^2+(y-\tfrac12)^2$.

    \item \textbf{Boundary-preserving Radial Expansion}: We have
    \begin{equation}
    \label{eq:test_map_T4}
    \begin{aligned}
    u_4(x,y) &= x + a\,s(x,y)e^{-r^2/\sigma}\left(x-\tfrac12\right), \ 
    v_4(x,y) = y + a\,s(x,y)e^{-r^2/\sigma}\left(y-\tfrac12\right),
    \end{aligned}
    \end{equation}
    where $s(x,y)=\sin(\pi x)\sin(\pi y)$,
    $r^2=(x-\tfrac12)^2+(y-\tfrac12)^2$,
    $a=0.55$, and $\sigma=0.10$.
\end{enumerate}

\begin{figure}[t!]
    \centering
    \includegraphics[width=\linewidth]{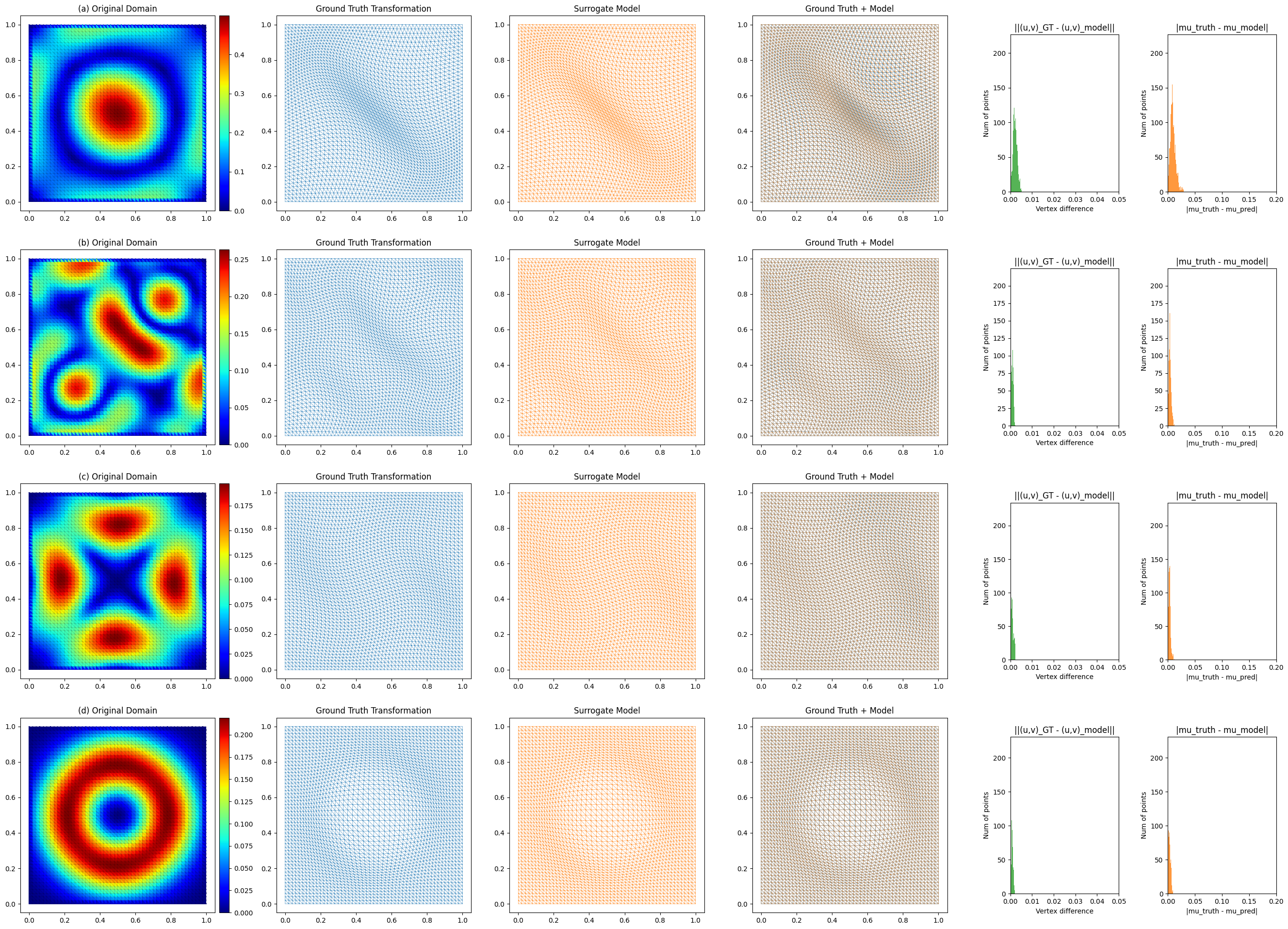}
    \caption{\textbf{Results of the reconstruction from different Beltrami coefficients.} Each row corresponds to one transformation induced by a different Beltrami coefficient:
    (a)~Eq.~\eqref{eq:test_map_T1},
    (b)~Eq.~\eqref{eq:test_map_T2},
    (c)~Eq.~\eqref{eq:test_map_T3},
    and (d)~Eq.~\eqref{eq:test_map_T4}. From left to right: Original domain with the Beltrami coefficients visualization, Ground truth we want to reconstruct, Reconstruction result by our surrogate model, Comparison of the truth and our prediction, Histogram of point difference value $|(u,v)_{\text{ref}}-(u,v)_{\text{pred}}|$ for our model, Histogram of value $|\mu_{\text{ref}}-\mu_{\text{pred}}|$ for our model.}
    \label{fig:bc_eq_mu}
\end{figure}

We evaluate reconstruction on test Beltrami coefficients induced by the analytic maps above. Fig.~\ref{fig:bc_eq_mu} compares the surrogate reconstructions with the analytic references and shows histograms of their discrepancies. Table~\ref{tab:reconstruction_results_mu} reports errors and runtimes for the surrogate.

\begin{table}[t]
\centering
\small
\begin{tabular}{c c c c}
\hline
\textbf{Test Case}  & \textbf{Time (s)} & \textbf{mean$(\bm{|\Delta d|})$} & \textbf{mean$(\bm{|\Delta\mu|})$} \\
\hline
(a) Eq.~\eqref{eq:test_map_T1}& 0.183517 & 0.00943353 & 0.01661039 \\
(b) Eq.~\eqref{eq:test_map_T2}& 0.178444 & 0.00495606 & 0.00628607 \\
(c) Eq.~\eqref{eq:test_map_T3}& 0.181272 & 0.00333415 & 0.00763160 \\
(d) Eq.~\eqref{eq:test_map_T4}& 0.178142 & 0.00323314 & 0.00575289 \\
\hline
\end{tabular}
\caption{\textbf{Reconstruction results under different Beltrami coefficients.} We report the computational time and reconstruction accuracy of the proposed model. The error is quantified by the mean absolute vertex-wise displacement difference between the ground truth and prediction, defined as $|\Delta d| = |(u, v)_{ref} - (u, v)_{pred}|$, together with the discrepancy in the Beltrami coefficient, $|\Delta \mu| = |\mu_{\text{ref}} - \mu_{\text{pred}}|$.}
\label{tab:reconstruction_results_mu}
\end{table}

\begin{figure}[t!]
    \centering
    \includegraphics[width=\linewidth]{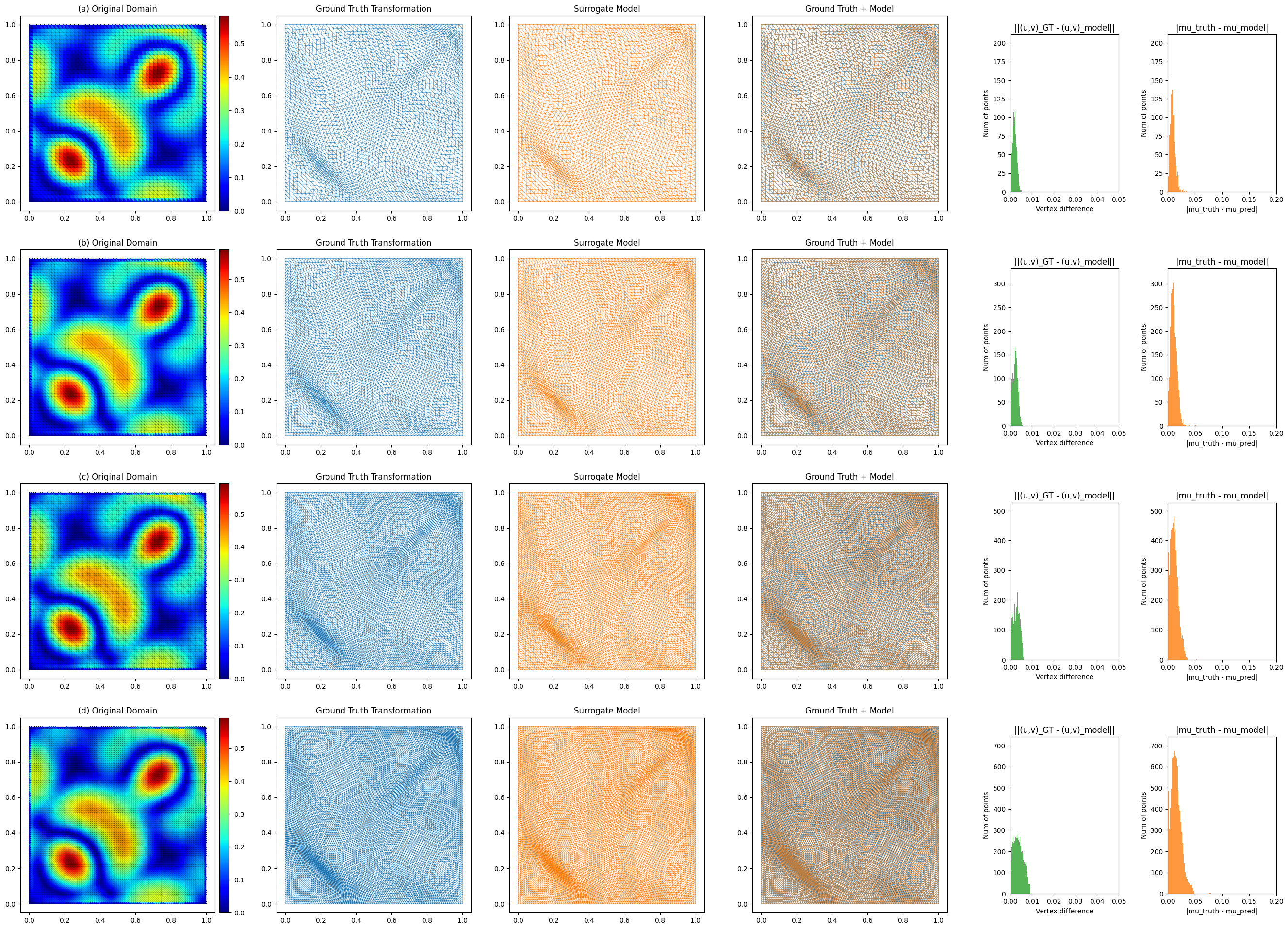}
    \caption{\textbf{Results of the reconstruction from different resolutions.} From left to right: Original domain with the Beltrami coefficients visualization, Ground truth we want to reconstruct, Reconstruction result by our surrogate model, Comparison of the truth and our prediction, Histogram of the point difference value $|(u,v)_{\text{truth}}-(u,v)_{\text{pred}}|$ for our model, Histogram of value $|\mu_{\text{truth}}-\mu_{\text{pred}}|$ for our model.}
    \label{fig:bc_eq_res}
\end{figure}

Furthermore, we evaluate the resolution-free characteristics of the framework by executing the mapping across various grid structures without any retraining. In this experiment, we use the Beltrami coefficient induced by the mild sinusoidal perturbation map in Eq.~\eqref{eq:test_map_T1} with $a=0.06, b=0.04$ as the test case. Fig.~\ref{fig:bc_eq_res} demonstrates the visual consistency maintained across these multi-resolution domains. The corresponding structural error metrics and computational scalability across the varying grid sizes are detailed in Table~\ref{tab:reconstruction_results_res}.

\begin{table}[t]
\centering
\small
\begin{tabular}{c c c c}
\hline
\textbf{$\bm{N}$} & \textbf{Time (s)} & \textbf{mean$(|\bm{\Delta d|)}$} & \textbf{mean$(|\bm{\Delta \mu|)}$} \\
\hline
48 & 0.224661 & 0.00191878 & 0.02008131 \\
64 & 0.250878 & 0.00226898 & 0.01994414 \\
80 & 0.246121 & 0.00282290 & 0.02708278 \\
96 & 0.283674 & 0.00380885 & 0.03982333 \\
\hline
\end{tabular}
\caption{\textbf{Reconstruction results for different resolutions.} We report the computational time and reconstruction accuracy of the proposed model. The error is quantified by the mean absolute vertex-wise displacement difference between the ground truth and prediction, defined as $|\Delta d| = |(u, v)_{ref} - (u, v)_{pred}|$, together with the discrepancy in the Beltrami coefficient, $|\Delta \mu| = |\mu_{\text{ref}} - \mu_{\text{pred}}|$.}
\label{tab:reconstruction_results_res}
\end{table}

From $N=48$ to $N=96$, the mean displacement error increases from $0.00192$ to $0.00381$ and the mean Beltrami error increases from $0.02008$ to $0.03982$, with a small decrease in the latter at $N=64$. Possible contributors include the smaller physical support of fixed convolution stencils on finer grids and the dependence of interpolation and derivative estimates on grid spacing. These results demonstrate reuse of the same trained model at the four tested grid sizes.

\subsection{DEM \& DEQ}

Density-equalizing mapping problems arise in a class of geometric variational problems that seek a mapping $T:\Omega\rightarrow\Omega$ whose local area distortion matches a prescribed population density $p(x,y)$. Given a density function defined on a planar domain, a density-equalizing map (DEM) redistributes area so that the pushforward density becomes
approximately uniform. Equivalently, the density-equalization with quasi-conformal constraints (DEQ) problem can be viewed as finding a deformation that balances the prescribed density while maintaining geometric regularity, typically enforced through distortion or quasi-conformality constraints. Such formulations play a fundamental role in cartograms, geometric processing, and variational mapping problems.

\subsubsection{2D DEQ \& DEM cases}
\label{subsubsec:2d_deq_dem}
We first evaluated our model on 2D DEQ and DEM problems. 

At each training iteration, a continuous population distribution $p(x,y)$ is synthesized on the unit square domain $\Omega = [0,1]^2$ as a randomized superposition of sinusoidal modes,

\begin{equation}
\label{eq:train_population}
p(x,y)
=
\sum_{i=1}^{n}
A_i
\sin \left(2\pi f_{x,i} x + \phi_{x,i}\right)
\cos\left(2\pi f_{y,i} y + \phi_{y,i}\right),
\end{equation}
where the number of modes $n$ is sampled uniformly from $[1,50]$. The amplitudes $A_i \in [-1,1]$, spatial frequencies $f_{x,i}, f_{y,i}$ are sampled log-uniformly from $[0.1,8.0]$, and the phase terms $\phi_{x,i}, \phi_{y,i}$ are sampled uniformly from $[0,2\pi]$.

After synthesis, the density $p(x,y)$ is shifted to ensure strict positivity and normalized to unit mass,
\[
\int_{\Omega} p(x,y)\,dx\,dy = 1.
\]

To enhance robustness and improve sampling efficiency, we adopt stochastic jittered sampling based on a Sobol quasi-random sequence. For each training epoch, a grid of resolution $N_{\mathrm{train}} \in [64,72]$ is generated, and the sampling coordinates are perturbed by an adaptively scaled jitter proportional to the local normalized gradient magnitude of $p(x,y)$. This strategy allocates denser samples to regions with large density variations and improves training stability.

The neural surrogate is trained using the following configuration:
\begin{itemize}
    \item \textbf{Epochs:} $12000$
    \item \textbf{Optimizer:} AdamW with learning rate $1\times 10^{-5}$ and weight decay $1\times 10^{-5}$
    \item \textbf{Learning-rate schedule:} Step decay with factor $0.5$ every $200$ epochs
    \item \textbf{Training grid resolution:} One uniform $N_{\mathrm{train}}\times N_{\mathrm{train}}$ grid per update, with $N_{\mathrm{train}}$ sampled uniformly from the integers $64$ through $96$
    \item \textbf{Network input:} the three-channel field $(x,\, y,\, p)$
    \item \textbf{Beltrami regularization weight:} $\lambda_{\mathrm{hr}} = 0.1$
    \item \textbf{Gradient clipping:} $\ell^2$-norm clipped at $1.0$
    \item \textbf{Hardware:} Single NVIDIA RTX~A6000 GPU; cuDNN disabled
\end{itemize}

For DEQ, the training loss is $\mathcal{L}(u,v)=\frac{1}{m}\sum_{i=1}^{m}L\!\left(\frac{p_i}{A_i+10^{-8}}-\frac{p}{A}\right)+\lambda_{\mathrm{hr}}\mathcal R_{\mathrm B}(u,v)$,
where $L(r)=\begin{cases}\frac12 r^2,& |r|<1,\\ |r|-\frac12,& |r|\geq1\end{cases}$ is the Smooth L1 loss and $m$ is the number of triangles. For triangle $i$, $p_i$ is its mass (the mean nodal population times the reference area) and $A_i$ is its mapped area. Here $p=h^2\sum_{k,l}\widehat p_{kl}$ is the discrete total population after nodal normalization and $A=1$ is the target area of the unit square. The Beltrami regularizer $\mathcal R_{\mathrm B}$ has weight $\lambda_{\mathrm{hr}}=0.1$ for DEQ. The DEM loss uses only the density term.

For the square-boundary case, $\mathcal B(\mathbf x)=\mathbf x\odot(\mathbf1-\mathbf x)$. For the free-boundary case, $\mathcal B(\mathbf x)=\mathbf1$. Directional refinement scales are applied separately as in Section~\ref{subsect:weight_refinement}.

To evaluate generalization across diverse population patterns, we set the refinement objective to be
\[
\mathcal{J}=\frac{1}{m}\sum_{i=1}^{m}L\!\left(\frac{p_i}{A_i+10^{-8}}-\frac{p}{A}\right) 
+10^{3}\max\{\max_i|\mu_i|-1,\,0\},
\]
where $\mu_i$ is the Beltrami coefficient on triangle $i$. The trained model is tested on 4 representative population distributions with distinct functional parameters:
\begin{enumerate}[(a)]

    \item
    \textbf{Sinusoidal--Cosine Interaction}: We set
    \begin{equation}
    \label{eqn:sin-cos}
    p_{\mathrm{test}}^{(1)}(x,y)
    =
    2 + \sin(2\pi x)\cos(2\pi y),
    \end{equation}

    \item
    \textbf{Exponential--Logarithmic Modulation}: We set
    \begin{equation}
    \label{eqn:exp-log}
    p_{\mathrm{test}}^{(2)}(x,y)
    =
    2
    + \sin \left(2\pi e^{x}\right)
    \cos \left(\pi \log(2y+0.1)\right),
    \end{equation}

    \item 
    \textbf{Rectangular Peak}: We set
    \begin{equation}
    \label{eqn:rect-peak}
    p_{\mathrm{test}}^{(3)}(x,y)
    =
    p_{\mathrm{test}}^{(1)}(x,y)
    +
    \operatorname{rect}(x,y),
    \end{equation}
    where $p_{\mathrm{test}}^{(1)}(x,y)$ is the sinusoidal--cosine background defined in
    Eq.~\eqref{eqn:sin-cos}, and $\operatorname{rect}(x,y)$ denotes a compactly supported rectangular peak.

    \item 
    \textbf{Ring-Shaped Density}: We set
    \begin{equation}
    \label{eqn:ring}
    p_{\mathrm{test}}^{(4)}(x,y)
    =
    \exp\!\left(
    -\frac{
    \left(\sqrt{(x-0.5)^2+(y-0.5)^2}-1.0\right)^2
    }{2(0.5)^2}
    \right),
    \end{equation}

\end{enumerate}

The four population test cases are evaluated on a $51\times51$ grid. Quantitative evaluation includes the standard deviation of the resulting density, the signed Jacobian determinant on triangular elements, and the spatial distribution of the Beltrami coefficient magnitude on the mapped domain.

For each $N\times N$ test grid with spacing $h=1/(N-1)$, let $p_{ij}$ be the sampled raw population. We shift the samples to have a minimum of $0.1$ and normalize by the discrete mass,
\[
p^+_{ij}=p_{ij}-\min_{k,l}p_{kl}+0.1,
\qquad
\widehat p_{ij}=\frac{p^+_{ij}}{h^2\sum_{k,l}p^+_{kl}}.
\]

\begin{figure}[t!]
    \centering
    \includegraphics[width=\linewidth]{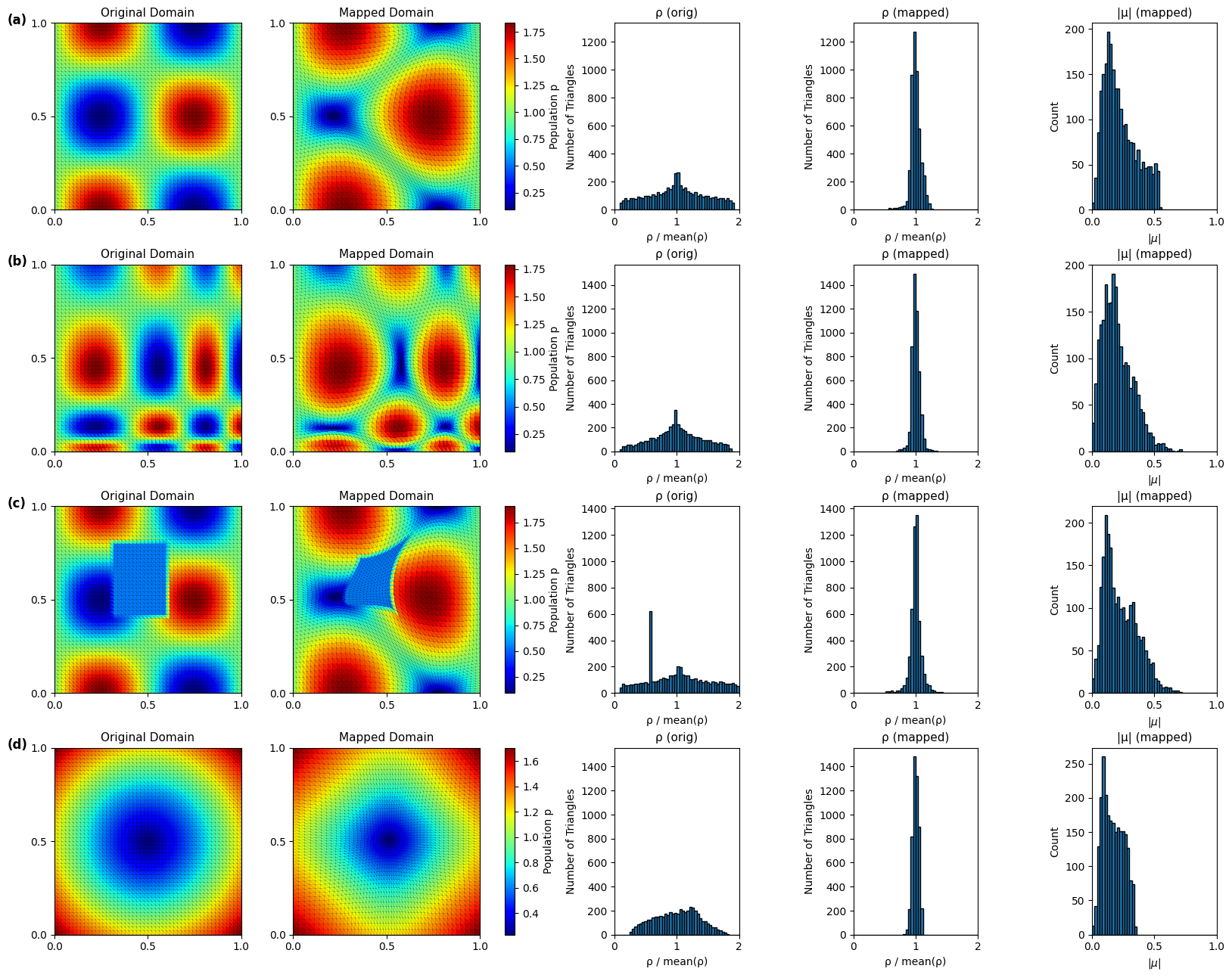}
    \caption{\textbf{2D DEQ population test results under square boundary condition.}
    Each row corresponds to one test distribution:
    (a) Sin--Cos (Eq.~\eqref{eqn:sin-cos}),
    (b) Exp--Log (Eq.~\eqref{eqn:exp-log}),
    (c) Rectangle Peak (Eq.~\eqref{eqn:rect-peak}),
    and (d) Ring (Eq.~\eqref{eqn:ring}).
    From left to right: original domain visualization, mapped domain
    visualization, density on the original domain, density on the mapped
    domain, and the magnitude of the Beltrami coefficient on the mapped domain.}
    \label{fig:population_test}
\end{figure}

\begin{table}[t]
\centering
\resizebox{\columnwidth}{!}{
\begin{tabular}{c c c c c c c}
\hline
\textbf{Test Case} & \textbf{Time (s)} & \textbf{Std$(\tilde{\bm{\rho}}_{orig})$} & \textbf{Std$(\tilde{\bm{\rho}}_{map})$} & \textbf{Min Jacobian} & \textbf{Max$(|\bm{\mu|)}$} & \textbf{Mean$(|\bm{\mu}|)$} \\
\hline
(a) Eq.~\eqref{eqn:sin-cos} & 0.804569 & 0.453812 & 0.082541 & 0.100071 & 0.544073 & 0.225081 \\
(b) Eq.~\eqref{eqn:exp-log} & 0.504526 & 0.397409 & 0.064182 & 0.085428 & 0.716671 & 0.203657 \\
(c) Eq.~\eqref{eqn:rect-peak} & 0.649909 & 0.480728 & 0.093411 & 0.099442 & 0.704816 & 0.228850 \\
(d) Eq.~\eqref{eqn:ring} & 0.312899 & 0.349701 & 0.048789 & 0.238288 & 0.346714 & 0.169448 \\
\hline
\end{tabular}
}
\caption{\textbf{The performance of our 2D DEQ model on different population cases under square boundary.} For each test case, we measured the time taken for the model execution, the standard deviation of the normalized density $\tilde{\rho} = \frac{\rho}{\text{Mean}(\rho)}$, where $\rho = \frac{\text{given population}}{\text{final area of each triangle}}$ on the original and mapped domains, the minimum signed triangle Jacobian determinant, the maximum and mean value of the norm of the Beltrami coefficient $\mu$.}
\label{tab:population_test}
\end{table}

\begin{figure}[t!]
    \centering
    \includegraphics[width=\linewidth]{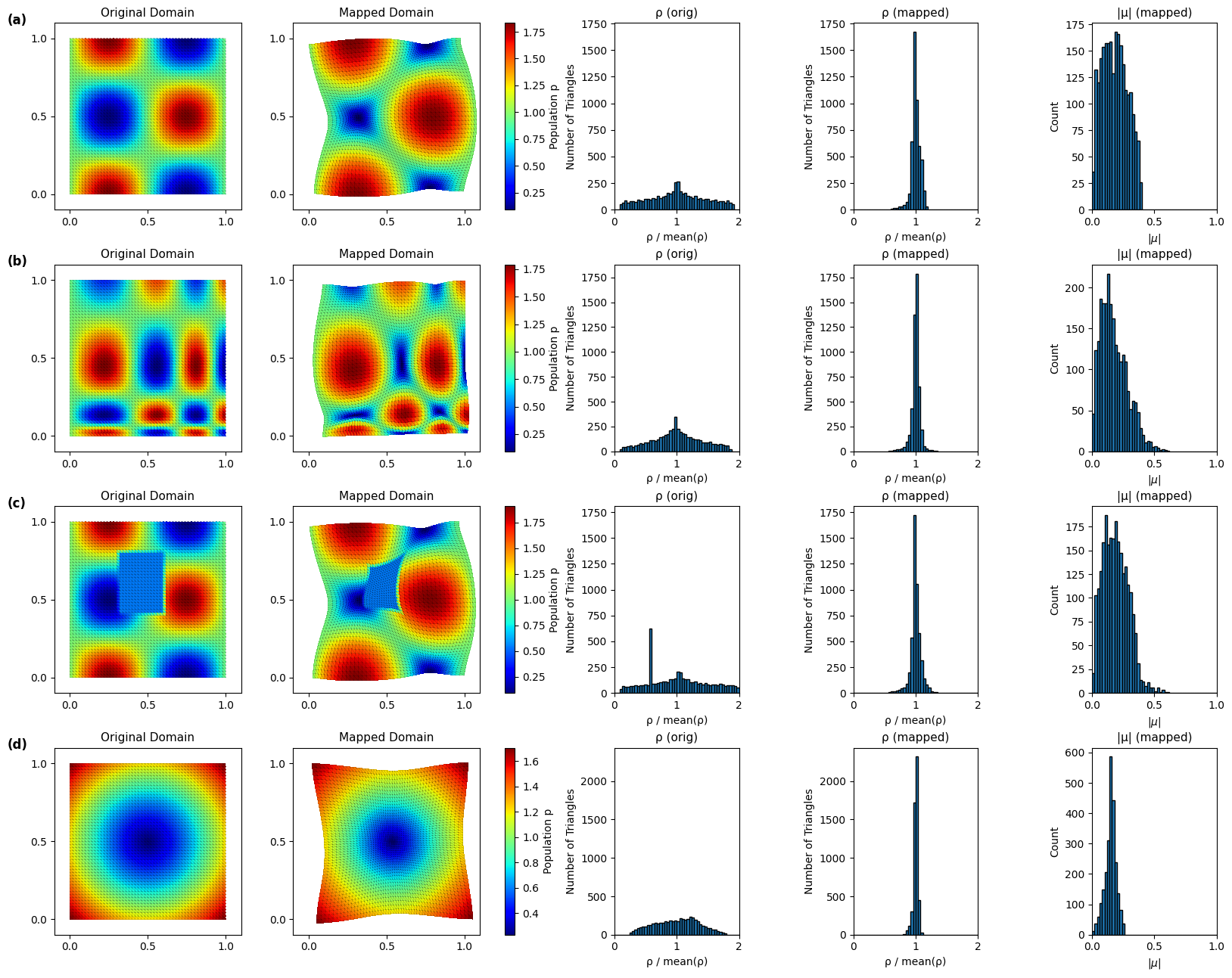}
    \caption{\textbf{2D DEQ population test results with free boundary.} Each row corresponds to one test distribution:
    (a) Sin--Cos (Eq.~\eqref{eqn:sin-cos}),
    (b) Exp--Log (Eq.~\eqref{eqn:exp-log},
    (c) Rectangle Peak (Eq.~\eqref{eqn:rect-peak}),
    and (d) Ring (Eq.~\eqref{eqn:ring}).
    From left to right: original domain visualization, mapped domain
    visualization, density on the original domain, density on the mapped
    domain, and the magnitude of the Beltrami coefficient on the mapped domain.}
    \label{fig:population_test_fb}
\end{figure}

To demonstrate the framework's flexibility to alternative boundary conditions, the four test configurations are similarly evaluated under a free boundary setup. Fig.~\ref{fig:population_test_fb} displays the resulting unconstrained spatial deformations, while Table~\ref{tab:population_test_fb} reports the corresponding tracking metrics and Jacobian values under this open boundary formulation.

\begin{table}[t]
\centering
\resizebox{\columnwidth}{!}{
\begin{tabular}{c c c c c c c}
\hline
\textbf{Test Case} & \textbf{Time (s)} & \textbf{Std$(\tilde{\bm{\rho}}_{orig})$} & \textbf{Std$(\tilde{\bm{\rho}}_{map})$} & \textbf{Min Jacobian} & \textbf{Max$(|\bm{\mu|)}$} & \textbf{Mean$(|\bm{\mu}|)$} \\
\hline
(a) Eq.~\eqref{eqn:sin-cos} & 0.251685 & 0.453812 & 0.072977 & 0.1117707 & 0.393088 & 0.183440 \\
(b) Eq.~\eqref{eqn:exp-log} & 0.210106 & 0.397409 & 0.070202 & 0.122339 & 0.605658 & 0.175640 \\
(c) Eq.~\eqref{eqn:rect-peak} & 0.214805 & 0.480728 & 0.087137 & 0.121764 & 0.601496 & 0.188528 \\
(d) Eq.~\eqref{eqn:ring} & 0.206635 & 0.349701 & 0.036166 & 0.270623 & 0.259938 & 0.147135 \\
\hline
\end{tabular}
}
\caption{\textbf{The performance of our 2D DEQ model on different population cases with free boundary.} For each test case, we measured the time taken for the model execution, the standard deviation of the normalized density $\tilde{\rho} = \frac{\rho}{\text{Mean}(\rho)}$, where $\rho = \frac{\text{given population}}{\text{final area of each triangle}}$ on the original and mapped domains, the minimum signed triangle Jacobian determinant, the maximum and mean value of the norm of the Beltrami coefficient $\mu$.}
\label{tab:population_test_fb}
\end{table}

The results indicate that the proposed method produces mappings that are nearly density-equalized across all tested population distributions while maintaining a small Beltrami distortion for both square boundary and free boundary cases.

To further assess the resolution-independent behavior of the proposed framework, we evaluate the same trained model on grids of varying resolutions $N\times N$ with $N \in \{32, 64, 128, 256\}$, using the multi-frequency test distribution in
\textbf{Multi-Frequency Composition}
\begin{equation}
\label{exp:res_test_f}
    p_{\mathrm{test}}(x,y)
    =
    2 + \sin(4\pi x)\cos(4\pi y) + 0.3\sin(6\pi x),
\end{equation}
shifted and normalized by the same discrete procedure, without retraining or architectural modification.

\begin{figure}[t!]
    \centering
    \includegraphics[width=\linewidth]{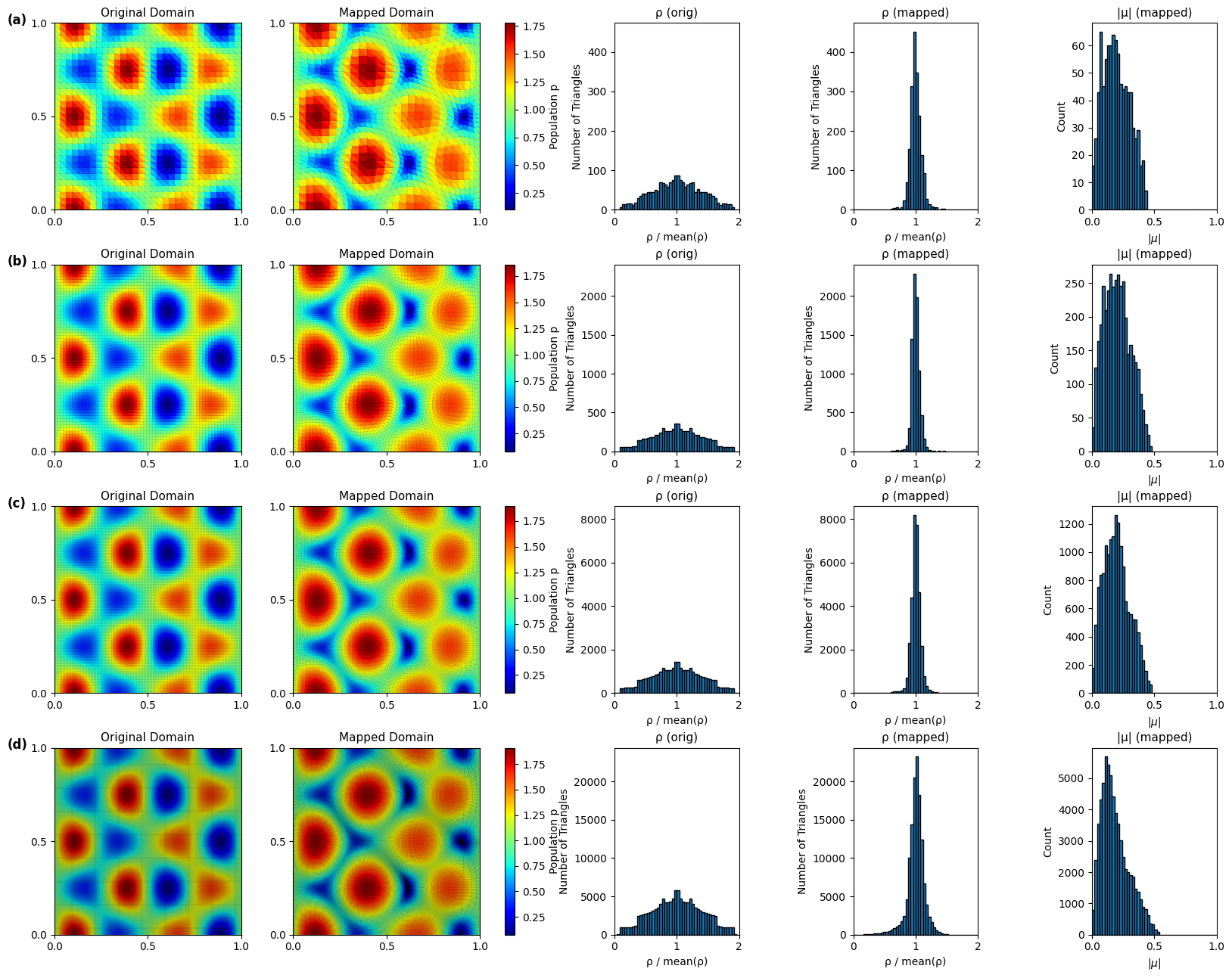}
    \caption{\textbf{2D DEQ resolution test results under square boundary condition.}
    Each row corresponds to a different grid resolution:
    (a) $N=32$, (b) $N=64$, (c) $N=128$,  and (d) $N=256$.
    From left to right: original domain, mapped domain, original density,
    mapped density, and Beltrami coefficient magnitude.}
    \label{fig:resolution_test}
\end{figure}

The framework's structural consistency across scales is highlighted in Fig.~\ref{fig:resolution_test}, which showcases stable deformation mappings from low to high point densities. To support these visual results, Table~\ref{tab:resolution_test} provides a quantitative report on the stability of the density metrics and structural variables across the evaluated resolutions.

\begin{table}[t]
\centering
\small
\begin{tabular}{c c c c c c c}
\hline
\textbf{$\bm{N}$} & \textbf{Time (s)} & \textbf{Std$(\tilde{\bm{\rho}}_{orig})$} & \textbf{Std$(\tilde{\bm{\rho}}_{map})$} & \textbf{Min Jacobian} & \textbf{Max$(|\bm{\mu|)}$} & \textbf{Mean$(|\bm{\mu}|)$} \\
\hline
32  & 0.467233 & 0.391159 & 0.090291 & 0.155488 & 0.438842 & 0.195081 \\
64  & 0.367197 & 0.400335 & 0.067485 & 0.075382 & 0.478099 & 0.202592 \\
128 & 0.458393 & 0.402907 & 0.078921 & 0.073391 & 0.476867 & 0.197402 \\
256 & 0.407083 & 0.403503 & 0.134581 & 0.118330 & 0.534828 & 0.184581 \\
\hline
\end{tabular}
\caption{\textbf{The performance of our 2D DEQ model on different resolution cases under square boundary.} For each test case, we measured the time taken for the model execution, the standard deviation of the normalized density $\tilde{\rho} = \frac{\rho}{\text{Mean}(\rho)}$, where $\rho = \frac{\text{given population}}{\text{final area of each triangle}}$ on the original and mapped domains, the minimum signed triangle Jacobian determinant, the maximum and mean value of the norm of the Beltrami coefficient $\mu$.}
\label{tab:resolution_test}
\end{table}
These experiments confirm that the learned mapping remains stable across resolutions, demonstrating the resolution-free nature of the proposed neural surrogate for density-equalizing maps in the 2D cases.

\begin{figure}[t!]
    \centering
    \includegraphics[width=\linewidth]{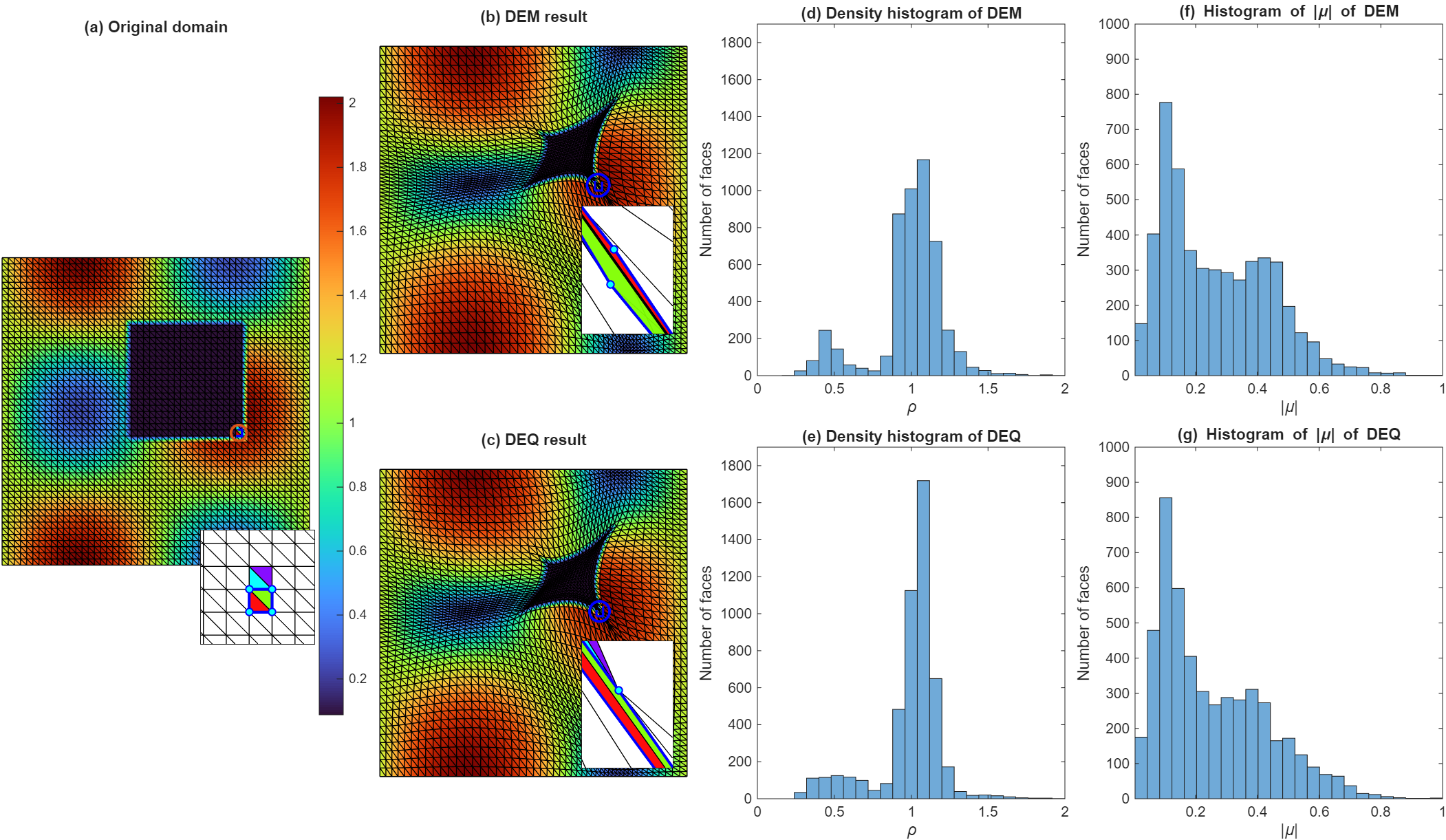}
    \caption{\textbf{Comparison between 2D DEM model and DEQ model on extreme case.} (a)~The original domain. (b) The mapped domain obtained by the 2D DEM model. (c) The mapped domain obtained by the 2D DEQ model. (d) The final density histogram of the DEM model. (e) The final density histogram of the DEQ model. (f) The histogram of $|\mu|$ of the DEM model. (g) The histogram of $|\mu|$ of the DEQ model.
    }
    \label{fig:extreme_comparison}
\end{figure}

Finally, we compared the DEQ and DEM results to see how the quasi-conformal term implemented by the Beltrami coefficient could improve the robustness of the mapping result. The comparison result is shown in Fig.~\ref{fig:extreme_comparison}.

\subsubsection{3D DEM cases}
\label{subsubsec:3d_dem}
We also evaluated our method on 3D DEM problems to show the ability to handle large-scale problems. For 3D problems, we only need to replace the 2D operations in our model with the 3D version, such as changing the 2D convolution to 3D convolution. There is no more structure to change when applying to 3D problems.

For each training iteration, we generated a continuous population distribution $p(x,y,z)$ on the unit cube domain $\Omega = [0,1]^3$ as a randomized superposition of sinusoidal functions,
\begin{equation}
    p(x,y,z) = \sum_{i=1}^n A_i\sin(2\pi f_{x,i}x + \phi_{x,i})\cos(2\pi f_{y,i}y + \phi_{y,i})\sin(2\pi f_{z,i}z + \phi_{z,i}),
\end{equation}

where the number of modes $n$ is sampled uniformly from $[4,40]$. The amplitudes $A_i \in [-1,1]$, spatial frequencies $f_{x,i}, f_{y,i}, f_{z,i}$ are sampled log-uniformly from $[0.1,5.0]$, and the phase terms $\phi_{x,i}, \phi_{y,i}, \phi_{z,i}$ are sampled uniformly from $[0,2\pi]$.

After synthesis, the density $p(x,y,z)$ is shifted to ensure strict positivity and normalized to unit mass,
\[
\int_{\Omega} p(x,y,z)\,dx\,dy\,dz = 1.
\]

We trained the surrogate model using the following configuration:

\begin{itemize}
    \item \textbf{Epochs:} $4000$
    \item \textbf{Optimizer:} AdamW with learning rate $2\times 10^{-4}$, betas $(0.9,\,0.995)$, and weight decay $5\times 10^{-5}$
    \item \textbf{Learning-rate schedule:} Cosine annealing with warm restarts, initial restart period $T_0=600$, multiplier $T_{\text{mult}}=2$, and minimum learning rate $2\times 10^{-6}$
    \item \textbf{Training grid resolution:} One uniform $N_{\mathrm{train}}\times N_{\mathrm{train}}\times N_{\mathrm{train}}$ grid per update, with $N_{\mathrm{train}}$ sampled uniformly from the integers $32$ through $48$
    \item \textbf{Network input:} the four-channel field $(x,\, y,\, z,\, p)$
    \item \textbf{Gradient clipping:} $\ell^2$-norm clipped at $1.0$
    \item \textbf{Hardware:} Single NVIDIA RTX~A6000 GPU; cuDNN disabled
\end{itemize}

For the 3D DEM training loss, the contribution of tetrahedron $i$ is $\mathcal L_i(u,v,w)=L\!\left(\frac{p_i}{T_i}-\frac{p}{T}\right)$,
where $L$ is the Smooth L1 loss defined in Section~\ref{subsubsec:2d_deq_dem}, $p_i$ is the mass assigned to tetrahedron $i$, $T_i$ is its mapped volume, $p$ is the total population, and $T$ is the target volume.
The overall density loss is $\mathcal L(u,v,w)=\frac{1}{m}\sum_{i=1}^{m}\mathcal L_i(u,v,w)$, where $m$ is the number of tetrahedra.

We use the boundary mask $\mathcal B(\mathbf x)=\,\mathbf x\odot(\mathbf1-\mathbf x)$ for the cube boundary and $\mathcal B(\mathbf x)=\mathbf1$ for the free boundary.

To evaluate generalization across diverse population patterns, we set the refinement objective to be
\[
\mathcal{J}=\frac{1}{m}\sum_{i=1}^m L\!\left(\frac{p_i}{T_i}-\frac{p}{T}\right)
+10^{5}\max\{-\min_e\det J_e,\,0\}.
\]
The trained model is tested on 4 representative population distributions with distinct functional parameters:

\begin{enumerate}[(a)]

    \item \textbf{Sinusoidal--Cosine Interaction (3D)}: We have
    \begin{equation}
    \label{eqn:sin-cos-3d}
    p_{\mathrm{test}}^{(1)}(x,y,z)
    =
    2 + \sin(2\pi x)\cos(2\pi y)\sin(2\pi z),
    \end{equation}.

    \item \textbf{Spherical-Shell Density}: We have
    \begin{equation}
    \label{eqn:sp-3d}
    p_{\mathrm{test}}^{(2)}(x,y,z)
    =
    \max(p)
    -
    \exp\!\left(
    -\frac{\left(r(x,y,z)-1.0\right)^2}{2(0.5)^2}
    \right),
    \end{equation}
    with $r(x,y,z) = \sqrt{(x-0.5)^2+(y-0.5)^2+(z-0.5)^2}$, representing an inverted Gaussian shell centered at $(0.5,0.5,0.5)$.

    \item \textbf{Eight Octant Partition}: We first define a smooth transition function
    \begin{equation}
    \label{eqn:oct-3d-transition}
    s(t)
    =
    \frac{1}{1 + e^{-(t-0.5)/0.1}}.
    \end{equation}
    Then, we let $s_x=s(x), s_y=s(y), s_z=s(z)$ and define
    \begin{equation}
    \label{eqn:oct-3d}
    \begin{aligned}
    p_{\mathrm{test}}^{(3)}(x,y,z)
    &=
    1(1-s_x)(1-s_y)(1-s_z) +2s_x(1-s_y)(1-s_z) \\
    &\quad
    +3(1-s_x)s_y(1-s_z) +4s_xs_y(1-s_z) +5(1-s_x)(1-s_y)s_z\\
    &\quad
    +6s_x(1-s_y)s_z +7(1-s_x)s_ys_z +8s_xs_ys_z,
    \end{aligned}
    \end{equation}
    which smoothly blends eight octants with different density levels.

    \item \textbf{Sinusoidal Field with Rectangular Prism Peak}: We define
    \begin{equation}
    \label{eqn:peak-3d}
    p_{\mathrm{test}}^{(4)}(x,y,z)
    =
    \begin{cases}
    1, & (x,y,z)\in \mathcal Q,\\
    2 + \sin(4\pi x)\cos(2\pi y)\cos(2\pi z), & \text{otherwise},
    \end{cases}
    \end{equation}
    where $\mathcal Q = \left\{(x,y,z)\mid 0.6<x<1,\;0<y<0.4,\;0.4<z<0.8\right\}$ is a rectangular region.

\end{enumerate}

\begin{figure}[t!]
    \centering
    \includegraphics[width=\linewidth]{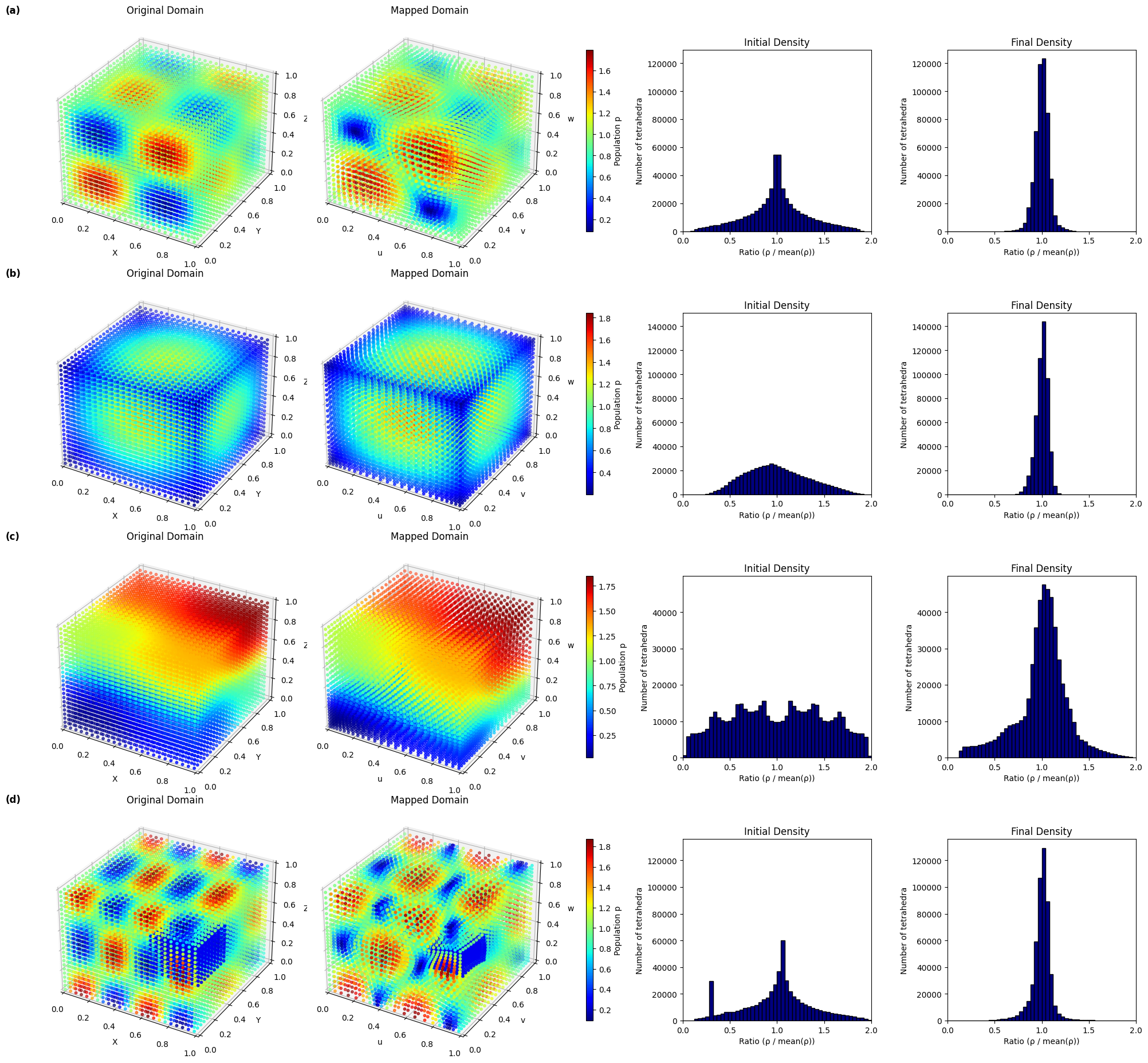}
    \caption{\textbf{3D DEM population test results under cube boundary condition.}
    Each row corresponds to one test distribution:
    (a) Sin--Cos (Eq.~\eqref{eqn:sin-cos-3d}),
    (b) Spherical (Eq.~\eqref{eqn:sp-3d}),
    (c) Eight octant (Eq.~\eqref{eqn:oct-3d}),
    and (d) Rectangular prism peak (Eq.~\eqref{eqn:peak-3d}).
    From left to right: original domain visualization, mapped domain visualization, density on the original domain, density on the mapped domain.}
    \label{fig:3d_population_test}
\end{figure}

We evaluate the four population cases on a $48\times48\times48$ grid. As in the 2D tests, we shift and normalize each raw test population before evaluation. We compute the density spread and signed Jacobian determinants on tetrahedral elements. Figure~\ref{fig:3d_population_test} shows the deformations with cube boundary, and Table~\ref{tab:3d_population_test} reports the runtimes, density statistics, and minimum signed Jacobian determinants.

The volumetric mapping behaviors for the 3D configurations within a standard rigid cube domain are visually presented in Fig.~\ref{fig:3d_population_test}. For a rigorous assessment, the corresponding numerical measurements evaluating volume preservation, convergence speed, and the minimum of the Jacobian determinants are summarized in Table~\ref{tab:3d_population_test}.

\begin{table}[t]
\centering
\small
\begin{tabular}{c c c c c}
\hline
\textbf{Test Case} & \textbf{Time (s)} & \textbf{Std$(\tilde{\bm{\rho}}_{orig})$} & \textbf{Std$(\tilde{\bm{\rho}}_{map})$} & \textbf{Min Jacobian} \\
\hline
(a) Eq.~\eqref{eqn:sin-cos-3d} & 0.758503 & 0.319948 & 0.073687 & 0.139083 \\
(b) Eq.~\eqref{eqn:sp-3d} & 0.764084 & 0.327913 & 0.063817 & 0.236052 \\
(c) Eq.~\eqref{eqn:oct-3d} & 0.755914 & 0.501315 & 0.275038 & 0.182880 \\
(d) Eq.~\eqref{eqn:peak-3d} & 0.755548 & 0.366671 & 0.110930 & 0.019724 \\
\hline
\end{tabular}
\caption{\textbf{The performance of our 3D DEM model on different population cases under cube boundary.} For each test case, we measured the time taken for the model execution, the standard deviation of the normalized density $\tilde{\rho} = \frac{\rho}{\text{Mean}(\rho)}$, where $\rho = \frac{\text{given population}}{\text{final volume of each tetrahedron}}$ on the original and mapped domains.}
\label{tab:3d_population_test}
\end{table}

\begin{figure}[t!]
    \centering
    \includegraphics[width=\linewidth]{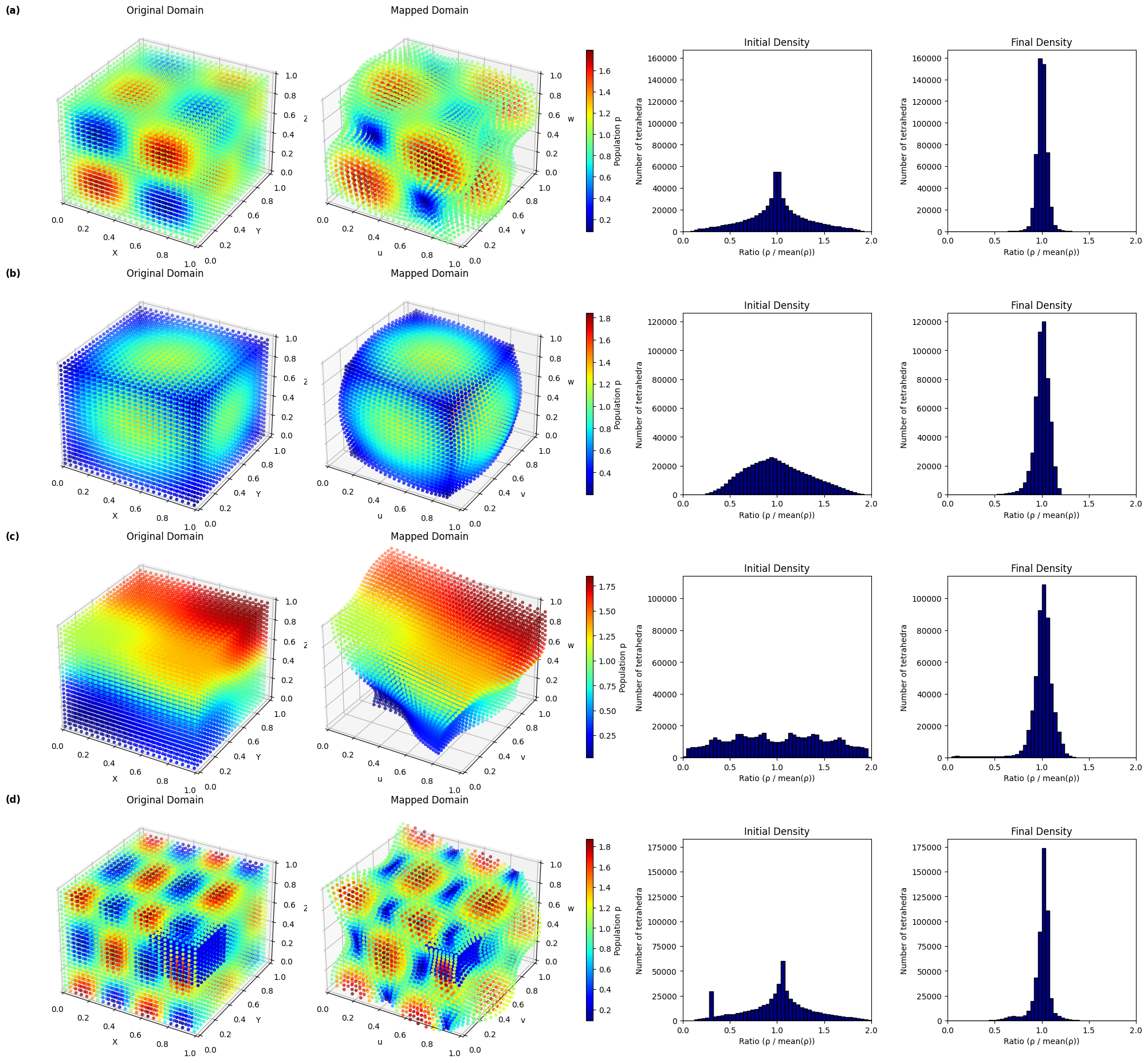}
    \caption{\textbf{3D DEM population test results with free boundary.} Each row corresponds to one test distribution:
    (a) Sin--Cos (Eq.~\eqref{eqn:sin-cos-3d}),
    (b) Spherical (Eq.~\eqref{eqn:sp-3d}),
    (c) Eight octant (Eq.~\eqref{eqn:oct-3d}),
    and (d) Rectangular prism peak (Eq.~\eqref{eqn:peak-3d}).
    From left to right: original domain visualization, mapped domain visualization, density on the original domain, density on the mapped domain.}
    \label{fig:3d_population_test_fb}
\end{figure}

To evaluate the model's performance under unconstrained boundary conditions in 3D scenarios, the computations are similarly executed within a free boundary environment. Fig.~\ref{fig:3d_population_test_fb} illustrates the qualitative volumetric deformations, while Table~\ref{tab:3d_population_test_fb} reports the corresponding density redistribution and geometric metrics under this free boundary formulation.

\begin{table}[t]
\centering
\small
\begin{tabular}{c c c c c}
\hline
\textbf{Test Case} & \textbf{Time (s)} & \textbf{Std$(\tilde{\bm{\rho}}_{orig})$} & \textbf{Std$(\tilde{\bm{\rho}}_{map})$} & \textbf{Min Jacobian} \\
\hline
(a) Eq.~\eqref{eqn:sin-cos-3d} & 0.991099 & 0.319948 & 0.055559 & 0.088221 \\
(b) Eq.~\eqref{eqn:sp-3d} & 0.980939 & 0.327913 & 0.079853 & 0.337495 \\
(c) Eq.~\eqref{eqn:oct-3d} & 0.952575 & 0.501315 & 0.137543 & 0.085793 \\
(d) Eq.~\eqref{eqn:peak-3d} & 0.944139 & 0.366671 & 0.100304 & 0.050543 \\
\hline
\end{tabular}
\caption{\textbf{The performance of our 3D DEM model on different population cases with free boundary.} For each test case, we measured the time taken for the model execution, the standard deviation of the normalized density $\tilde{\rho} = \frac{\rho}{\text{Mean}(\rho)}$, where $\rho = \frac{\text{given population}}{\text{final volume of each tetrahedron}}$ on the original and mapped domains.}
\label{tab:3d_population_test_fb}
\end{table}

To further assess the resolution-independent behavior of the proposed framework, we evaluate the same trained model on grids of varying resolutions
$N\times N$ with $N \in \{32, 40, 56, 64\}$, using the multi-frequency test distribution in
\textbf{Multi-Frequency Composition 3D}
\label{exp:res_test_f_3d}
    \[
        p_{\mathrm{test}}(x,y,z)
        =
        2 + \sin(4\pi x)\cos(2\pi y)\cos(2\pi z),
    \]
The population is normalized to unit mass before evaluation.

\begin{figure}[t!]
    \centering
    \includegraphics[width=\linewidth]{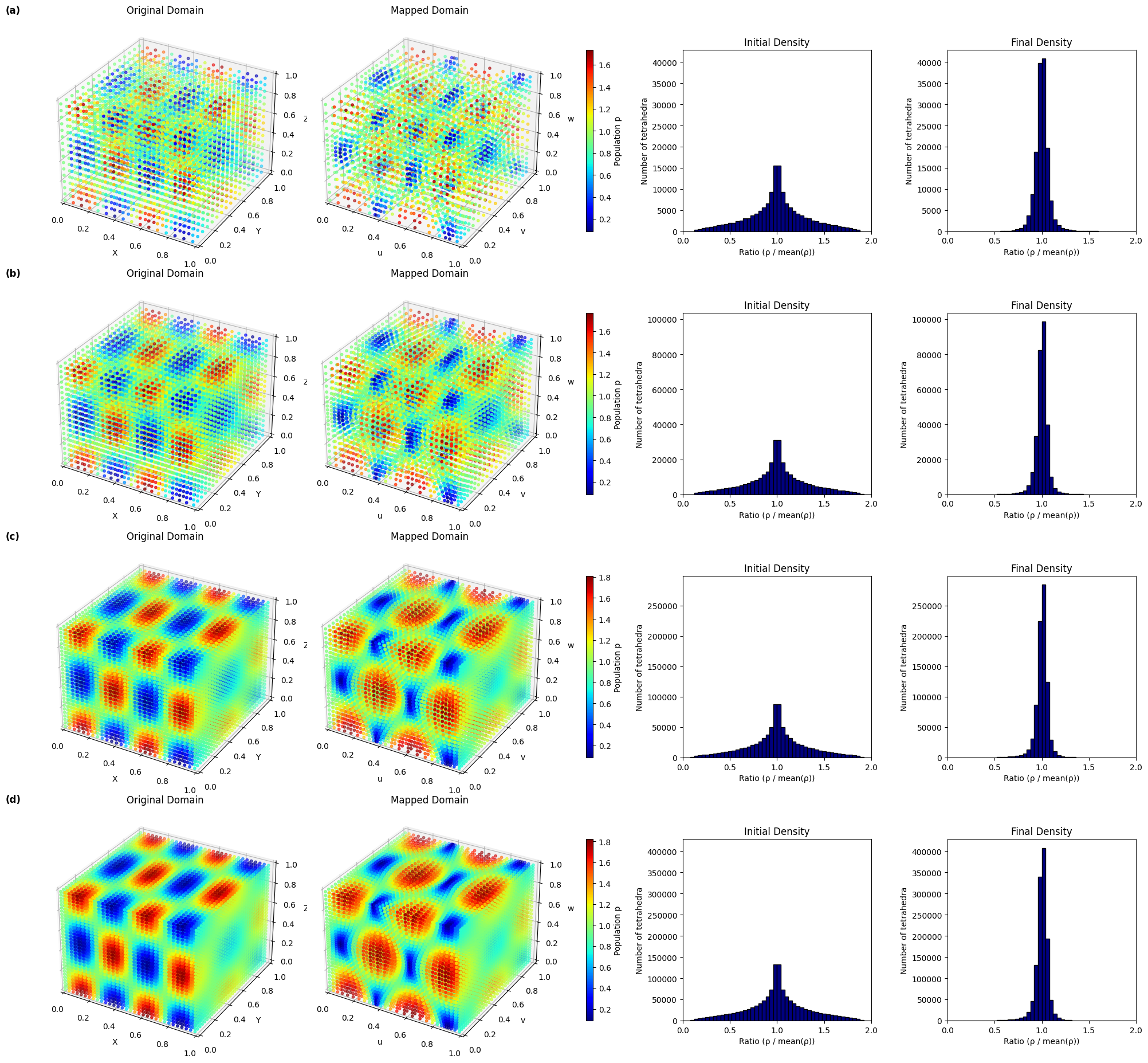}
    \caption{\textbf{3D DEM resolution test results under cube boundary condition.}
    Each row corresponds to a different grid resolution:
    (a) $N = 32$,
    (b) $N = 40$,
    (c) $N = 56$,
    and (d) $N = 64$.
    From left to right: original domain visualization, mapped domain
    visualization, density on the original domain, density on the mapped
    domain.}
    \label{fig:3d_resolution_test}
\end{figure}

Fig.~\ref{fig:3d_resolution_test} shows the 3D DEM predictions across different grid resolutions. To support these visual findings, the computational timings, density redistribution statistics, and minimum signed Jacobian determinants over all tetrahedra across the varying mesh scales are detailed in Table~\ref{tab:3d_resolution_test}.

For the cube-boundary experiment, the componentwise mask fixes the normal coordinate of each boundary vertex, so each boundary triangle lies in the supporting plane of its corresponding cube face. This assignment is topologically feasible because it is realized by the identity map. Since the minimum signed determinant is positive at every tested resolution, Theorem~2 of Lipman~\cite{lipman2014bijective} gives injectivity on the interior and shows that the image is the closed cube. Each boundary triangle has an adjacent positively oriented tetrahedron whose opposite vertex lies on the inward side of its face plane. This preserves the triangle's induced orientation. On cube edges, the same argument within each face gives the correct edge orientation, and the cube corners are fixed. Theorem~3 of Lipman~\cite{lipman2014bijective} then gives bijectivity on the closed cube. This conclusion concerns the reported piecewise-affine maps under the cube boundary mask.

\begin{table}[t]
\centering
\small
\begin{tabular}{c c c c c}
\hline
\textbf{$\bm{N}$} & \textbf{Time (s)} & \textbf{Std$(\tilde{\bm{\rho}}_{orig})$} & \textbf{Std$(\tilde{\bm{\rho}}_{map})$} & \textbf{Min Jacobian} \\
\hline
32 & 0.411478 & 0.313550 & 0.079741 & 0.061818 \\
40 & 0.594614 & 0.316422 & 0.073728 & 0.018812 \\
56 & 1.315721 & 0.318893 & 0.074921 & 0.000006 \\
64 & 3.077169 & 0.319491 & 0.075686 & 0.000028 \\
\hline
\end{tabular}
\caption{\textbf{The performance of our 3D DEM model on different resolution cases under cube boundary.} For each test case, we measured the time taken for the model execution, the standard deviation of the normalized density $\tilde{\rho} = \frac{\rho}{\text{Mean}(\rho)}$, where $\rho = \frac{\text{given population}}{\text{final volume of each tetrahedron}}$ on the original and mapped domains.}
\label{tab:3d_resolution_test}
\end{table}

\subsection{Comparison with LDEM and ablation of weight refinement}
We compare the DEM surrogate with LDEM (with fine-tuning) on the population and resolution cases of this section. Both methods use the same population, the same grid, and the square or cube boundary. We also set the training loss for both the LDEM method and our surrogate to be Eq. (4.15) for 2D and Eq. (7.7) for 3D proposed by the LDEM paper~\cite{huang2025ldem} to preserve consistency. LDEM optimizes a network for each new population. The surrogate evaluates one trained network with the same hyperparameters configuration in Section~\ref{subsubsec:2d_deq_dem} and~\ref{subsubsec:3d_dem}, then runs the scale search of Section~\ref{subsect:weight_refinement}. The surrogate is trained on a single NVIDIA RTX~A6000 GPU. Timings in Table~\ref{tab:ldem_compare} are wall-clock seconds on that GPU with cuDNN disabled. The LDEM paper evaluates on an Intel Xeon Gold 6238R CPU at 2.20\,GHz, and those CPU timings are reported in Table~2 of \cite{huang2025ldem}. A CPU run can be faster for simple cases, such as 2D grids at low resolution. LDEM optimizes a new network through many short sequential steps, and on a small grid each step does little arithmetic, so GPU kernel launch and host-device transfer costs can exceed the computation. On that CPU, every 3D case takes more than 10 minutes, so this table uses the GPU timings. Accuracy is the standard deviation of the normalized element density, and orientation is the minimum signed element Jacobian together with the number of elements whose Jacobian is negative.

On the four 2D populations, the surrogate takes $0.484$--$1.593$\,s and LDEM takes $24.402$--$38.168$\,s. LDEM attains lower mapped density spread in all four cases. On the 2D resolution series, LDEM attains lower spread at $N=32$, $64$, and $256$, and the surrogate attains lower spread at $N=128$. Both methods have positive minimum signed Jacobians at these sizes. On the four 3D populations, the surrogate takes $0.768$--$0.773$\,s and LDEM takes $84.042$--$86.915$\,s. The surrogate attains lower density spread in all four 3D cases. Its minimum signed Jacobian in the octant case is approximately $10^{-5}$, which is still a positive value, indicating the map is still bijective.

\begin{table}[t]
\centering
\resizebox{\linewidth}{!}{%
\begin{tabular}{@{}l|ccc|ccc@{}}
\hline
\textbf{Test Case} & \textbf{Ours (s)} & \textbf{Std map} & \textbf{Min Jacobian} & \textbf{LDEM (s)} & \textbf{Std map} & \textbf{Min Jacobian} \\
\hline
(a) 2D Eq.~\eqref{eqn:sin-cos}, $N=51$ & 1.593 & 0.085730 & 0.082558 & 24.402 & 0.014471 & 0.089175 \\
(b) 2D Eq.~\eqref{eqn:exp-log}, $N=51$ & 0.484 & 0.076477 & 0.120175 & 38.168 & 0.035255 & 0.099470 \\
(c) 2D Eq.~\eqref{eqn:rect-peak}, $N=51$ & 1.418 & 0.095039 & 0.088587 & 25.784 & 0.023972 & 0.092362 \\
(d) 2D Eq.~\eqref{eqn:ring}, $N=51$ & 0.634 & 0.054539 & 0.279841 & 25.398 & 0.009297 & 0.246911 \\
(e) 2D Eq.~\eqref{exp:res_test_f}, $N=32$ & 0.901 & 0.157882 & 0.094327 & 19.792 & 0.040597 & 0.106848 \\
(f) 2D Eq.~\eqref{exp:res_test_f}, $N=64$ & 0.716 & 0.062598 & 0.113597 & 22.847 & 0.022603 & 0.082717 \\
(g) 2D Eq.~\eqref{exp:res_test_f}, $N=128$ & 0.648 & 0.070183 & 0.100501 & 20.172 & 0.097715 & 0.073093 \\
(h) 2D Eq.~\eqref{exp:res_test_f}, $N=256$ & 1.204 & 0.133909 & 0.089832 & 24.301 & 0.110749 & 0.061452 \\
(i) 3D Eq.~\eqref{eqn:sin-cos-3d}, $N=48$ & 0.773 & 0.067233 & 0.091529 & 86.915 & 0.085185 & 0.117096 \\
(j) 3D Eq.~\eqref{eqn:sp-3d}, $N=48$ & 0.769 & 0.073342 & 0.210777 & 84.042 & 0.169159 & 0.426152 \\
(k) 3D Eq.~\eqref{eqn:oct-3d}, $N=48$ & 0.768 & 0.200275 & 0.000010 & 86.839 & 0.460561 & 0.745663 \\
(l) 3D Eq.~\eqref{eqn:peak-3d}, $N=48$ & 0.769 & 0.110694 & 0.042410 & 86.713 & 0.135059 & 0.091413 \\
\hline
\end{tabular}%
}
\caption{\textbf{DEM comparison with LDEM.}
Each row corresponds to one test case:
(a) 2D Eq.~\eqref{eqn:sin-cos}, $N=51$,
(b) 2D Eq.~\eqref{eqn:exp-log}, $N=51$,
(c) 2D Eq.~\eqref{eqn:rect-peak}, $N=51$,
(d) 2D Eq.~\eqref{eqn:ring}, $N=51$,
(e) 2D Eq.~\eqref{exp:res_test_f}, $N=32$,
(f) 2D Eq.~\eqref{exp:res_test_f}, $N=64$,
(g) 2D Eq.~\eqref{exp:res_test_f}, $N=128$,
(h) 2D Eq.~\eqref{exp:res_test_f}, $N=256$,
(i) 3D Eq.~\eqref{eqn:sin-cos-3d}, $N=48$,
(j) 3D Eq.~\eqref{eqn:sp-3d}, $N=48$,
(k) 3D Eq.~\eqref{eqn:oct-3d}, $N=48$,
and (l) 3D Eq.~\eqref{eqn:peak-3d}, $N=48$.
From left to right: test case, surrogate time, mapped density standard deviation, minimum Jacobian, LDEM time, LDEM mapped density standard deviation, and LDEM minimum Jacobian.}
\label{tab:ldem_compare}
\end{table}

For 2D DEQ, the refinement adds $10^3\max\{\max_i|\mu_i|-1,0\}$ to the density loss, where $\mu_i$ is the Beltrami coefficient on triangle $i$. The grid-node Beltrami estimates are computed separately. Table~\ref{tab:refine_ablation} shows that refinement reduces the mapped density spread on all four population cases and at $N=64$, $128$, and $256$. At $N=32$ the spread increases slightly from $0.088936$ to $0.090297$. At $N=256$ the raw map has $3663$ inverted triangles. The refined map has zero inversions and a minimum signed Jacobian of $0.118313$. The forward pass takes $0.025$--$0.052$\,s. The refinement takes $0.616$--$1.658$\,s. The refinement time stays short on every case, so this small amount of extra computation improves the mapping quality and preserves bijectivity.

\begin{table}[t]
\centering
\resizebox{\linewidth}{!}{%
\begin{tabular}{@{}lcccccccc@{}}
\hline
\textbf{Test Case} & \textbf{Raw std} & \textbf{Refined std} & \textbf{Raw min Jac.} & \textbf{Refined min Jac.} & \textbf{Raw inv.} & \textbf{Refined inv.} & \textbf{Model (s)} & \textbf{Refinement (s)} \\
\hline
(a) Eq.~\eqref{eqn:sin-cos}, $N=51$ & 0.127270 & 0.082534 & 0.184226 & 0.100095 & 0 & 0 & 0.026280 & 1.657622 \\
(b) Eq.~\eqref{eqn:exp-log}, $N=51$ & 0.076303 & 0.064181 & 0.131301 & 0.085400 & 0 & 0 & 0.025489 & 1.120086 \\
(c) Eq.~\eqref{eqn:rect-peak}, $N=51$ & 0.127019 & 0.093410 & 0.184626 & 0.099451 & 0 & 0 & 0.025502 & 1.367170 \\
(d) Eq.~\eqref{eqn:ring}, $N=51$ & 0.104689 & 0.048789 & 0.388873 & 0.238288 & 0 & 0 & 0.026075 & 0.616421 \\
(e) Eq.~\eqref{exp:res_test_f}, $N=32$ & 0.088936 & 0.090297 & 0.176304 & 0.155471 & 0 & 0 & 0.024711 & 1.373020 \\
(f) Eq.~\eqref{exp:res_test_f}, $N=64$ & 0.072187 & 0.067487 & 0.103420 & 0.075375 & 0 & 0 & 0.026207 & 1.118128 \\
(g) Eq.~\eqref{exp:res_test_f}, $N=128$ & 0.121927 & 0.078921 & 0.038039 & 0.073381 & 0 & 0 & 0.027274 & 1.390784 \\
(h) Eq.~\eqref{exp:res_test_f}, $N=256$ & 17.519581 & 0.134582 & $-0.184451$ & 0.118313 & 3663 & 0 & 0.052471 & 1.133497 \\
\hline
\end{tabular}%
}
\caption{\textbf{Ablation of weight refinement on the 2D DEQ task.}
Each row corresponds to one test case:
(a) Eq.~\eqref{eqn:sin-cos}, $N=51$,
(b) Eq.~\eqref{eqn:exp-log}, $N=51$,
(c) Eq.~\eqref{eqn:rect-peak}, $N=51$,
(d) Eq.~\eqref{eqn:ring}, $N=51$,
(e) Eq.~\eqref{exp:res_test_f}, $N=32$,
(f) Eq.~\eqref{exp:res_test_f}, $N=64$,
(g) Eq.~\eqref{exp:res_test_f}, $N=128$,
and (h) Eq.~\eqref{exp:res_test_f}, $N=256$.
From left to right: test case, raw mapped density standard deviation, refined mapped density standard deviation, raw minimum Jacobian, refined minimum Jacobian, raw inverted triangles, refined inverted triangles, model forward-pass time, and refinement time.}
\label{tab:refine_ablation}
\end{table}

\section{Conclusion}
\label{sect:conclusion}

In this paper, we presented a resolution-free neural surrogate approach designed to solve large-scale geometric parameterization and mapping problems with spatially varying parameter fields. By leveraging a structured-grid encoding that ingests multi-resolution samples of the coefficient field, our architecture reuses the same trained weights across the different grid sizes. The surrogate is trained in a data-free manner, utilizing physics-informed residuals and variational energy functionals to guide the optimization process without the need for expensive ground-truth labels.

Our experimental results on the reconstruction from Beltrami coefficients and density-equalizing maps (DEM/DEQ) demonstrate that the same trained model can be evaluated on multiple grid sizes in each task. Specifically, the surrogate achieves high accuracy in both 2D and 3D settings while maintaining stability across varying grid densities at inference time. The integration of an edge-aware convolution operator and a low-dimensional weight refinement step further ensures that the mapping remains geometrically consistent and bijectivity is preserved even in high-distortion scenarios.

The primary advantage of our approach lies in its efficiency; once trained, the model provides near real-time solutions for diverse parameter fields that would otherwise require iterative, computationally intensive numerical solvers. Future work could involve extending this framework to more complex geometric parameterization and mapping problems or even to more general geometric PDEs and variational problems~\cite{lai2013ridge}. Additionally, investigating hybrid architectures that combine the global spectral advantages of Fourier-based methods with our local resolution-free conditioning may further improve performance for high-frequency parameter fields.

\bibliographystyle{ieeetr}
\bibliography{reference.bib}

\end{document}